\documentclass{article} 
\usepackage{iclr2027_conference,times}
\usepackage{booktabs}
\usepackage{amssymb}
\usepackage{multirow}
\usepackage{makecell}
\usepackage{pifont}
\usepackage[table]{xcolor}

\definecolor{mycolor}{HTML}{a7caea}

\usepackage{amsmath,amsfonts,bm}

\def\1{\bm{1}}

\DeclareMathAlphabet{\mathsfit}{\encodingdefault}{\sfdefault}{m}{sl}
\SetMathAlphabet{\mathsfit}{bold}{\encodingdefault}{\sfdefault}{bx}{n}

\definecolor{linkcolor}{HTML}{ED1C24}
\definecolor{mydarkgreen}{rgb}{0.02,0.6,0.02}
\usepackage{algorithm}
\usepackage{listings}
\usepackage[colorlinks,citecolor=mydarkgreen,linkcolor=linkcolor]{hyperref}
\usepackage{url}
\usepackage{graphicx}
\usepackage{wrapfig}
\usepackage{capt-of}

\title{RealtimeWAM: One-Step Asynchronous World Action Models}

\author{\textbf{Chengtao Lv\textsuperscript{1}\thanks{Equal Contribution.}\quad
Jinyang Du\textsuperscript{2}\footnotemark[1]\quad
Shuyi Feng\textsuperscript{1}\quad
Yang Yong\textsuperscript{3}\quad
Shiqiao Gu\textsuperscript{3}}\\
\textbf{Shunzi Yang\textsuperscript{2}\quad
Ruihao Gong\textsuperscript{2}\thanks{Corresponding Authors.}\quad
Shen Ren\textsuperscript{4}\quad
Tianwei Zhang\textsuperscript{1}\quad
Wenya Wang\textsuperscript{1}\footnotemark[2]}\\
{\footnotesize\textsuperscript{1}Nanyang Technological University\quad
\textsuperscript{2}Beihang University\quad
\textsuperscript{3}Sensetime\quad
\textsuperscript{4}Continental Automotive Singapore}\\
{\footnotesize\texttt{chengtao001@e.ntu.edu.sg\quad wangwy@ntu.edu.sg}}\\
{\footnotesize\texttt{\{jinyangdu, gongruihao\}@buaa.edu.cn}}
}

\providecommand{\eg}{\emph{e.g.}}
\providecommand{\ie}{\emph{i.e.}}

\iclrfinalcopy 
\begin{document}

\maketitle
\lhead{Preprint}

\begin{abstract}

World Action Models (WAMs) incorporate visual representations from video generation backbones to guide action prediction. Recent efficient WAMs adopt Mixture-of-Transformers (MoT) architectures and compute video representations once for reuse by the action expert. However, intra-expert iteration (\ie, multi-step action denoising) and inter-expert waiting (\ie, sequential execution of the video and action experts) still limit inference efficiency. To this end, we present RealtimeWAM, an extremely efficient WAM variant with one-step action generation and asynchronous inference, addressing these two bottlenecks. To reduce intra-expert iteration, we propose Teacher-Anchored Consistency Distillation (TACD) to address a local-global error gap: low local consistency error alone does not guarantee accurate final actions. TACD supplements local consistency with explicit supervision from the frozen teacher's multi-step rollout endpoint, enabling accurate one-step action generation. Additionally, we propose Cross-Expert Wavefront Pipelining (CEWP) to eliminate unnecessary expert-level waiting. It overlaps the two experts through block-wise sharing of the video KV cache, synchronizing only immediately before the corresponding action attention consumes it. Extensive experiments across diverse benchmarks (\eg, LIBERO, LIBERO-Plus and RoboTwin) and model variants (\eg, Fast-WAM and Faster-WAM) demonstrate the superiority of RealtimeWAM. Notably, RealtimeWAM maintains near-lossless performance (\ie, $<1\%$ drop) across these benchmarks while delivering significant end-to-end speedup (\eg, $\sim25\times$ on H100).
Our code and checkpoints are available via this \href{https://github.com/ModelTC/LightX2V/tree/main/scripts/realtimewam}{link}.

\end{abstract}

\addtocontents{toc}{\protect\setcounter{tocdepth}{-1}}
\section{Introduction}
World Action Models (WAMs)~\citep{yuan2026fast,ye2026world,bi2026motus,li2026causal,kim2026cosmos,zhao2026faster} have emerged as a promising paradigm for robot manipulation by jointly modeling future visual observations and robot actions. Unlike conventional Vision-Language-Action (VLA) models~\citep{shukor2025smolvla,intelligence2025pi_,kim2024openvla,bjorck2025gr00t,black2024pi_0,liu2025rdt} that map visual observations and language instructions directly to actions, WAMs incorporate future visual prediction as an additional learning objective. This supervision encourages policies to capture the physical dynamics and temporal structure of robot--environment interactions.


Based on whether video and action modeling use the same network, WAM architectures can be broadly divided into two categories. Representative shared-backbone WAMs~\citep{kim2026cosmos,ye2026world} jointly denoise video and action tokens within a single backbone. Consequently, iterative denoising of long token sequences through the heavy video backbone increases inference latency. In contrast, some efficient Mixture-of-Transformers (MoT)-based WAMs~\citep{yuan2026fast,zhao2026faster} offer a promising alternative: the video expert computes representations \textit{once}, while a separate action expert reuses them for action generation.

\begin{figure}[!t]
    \centering
    \setlength{\abovecaptionskip}{0.2cm}
    \includegraphics[width=\linewidth]{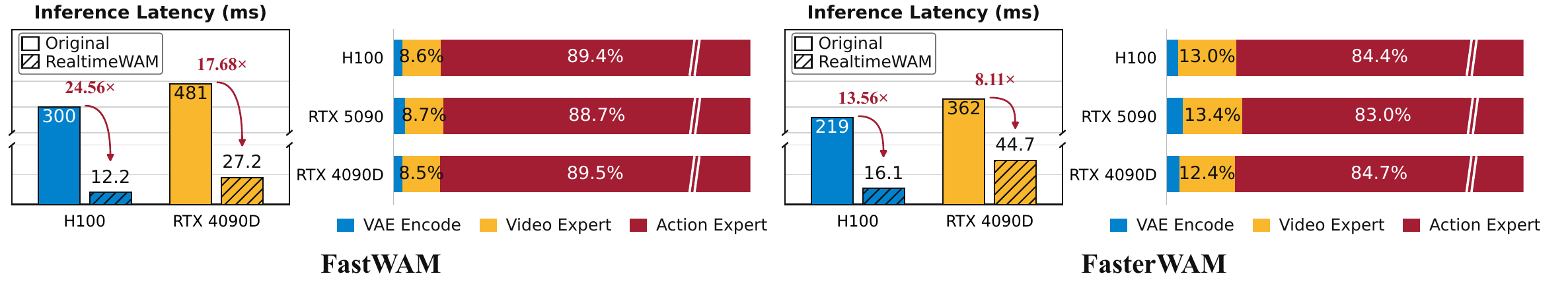}
    \caption{Inference efficiency of Fast-WAM (\textit{left}) and Faster-WAM (\textit{right}) across diverse hardware. Vertical bars compare inference latency between the original models and RealtimeWAM, while stacked bars show the runtime proportions of key components.}
    \label{fig:intro}
    \vspace{-0.1in}
\end{figure}

Despite advancements in inference pipeline design, these MoT-based WAMs still incur substantial latency. Specifically, even on advanced GPUs (\eg, H100 in Fig.~\ref{fig:intro}), these models (\eg, Fast-WAM~\citep{yuan2026fast}) require over 300~ms for a single inference call. We identify two inference bottlenecks that remain insufficiently addressed in existing MoT-based WAMs: \ding{192} \textit{Intra-expert iteration} arises from multi-step action denoising. We observe that the action expert accounts for nearly 90\% of end-to-end latency for action generation across diverse GPUs (see Fig.~\ref{fig:intro}). \ding{193} \textit{Inter-expert waiting} stems from sequential execution of the video and action experts. As shown in Fig.~\ref{fig:method2}(a), the action expert begins execution only after the video expert completes its forward pass, causing unnecessary waiting.

Motivated by the above analysis, we present \textit{RealtimeWAM}, an extremely efficient post-trained variant based on \textit{any} MoT-based WAM. To our knowledge, it is the \textit{first} one-step WAM to achieve near-lossless performance compared with its multi-step counterparts.

To combat intra-expert iteration, we propose \textit{Teacher-Anchored Consistency Distillation} (TACD) for one-step action generation. TACD addresses a \textit{local-global error gap} in conventional consistency distillation, which only constrains local consistency while neglecting to explicitly penalize the global target action error. TACD adopts the frozen teacher's multi-step rollout endpoint as a numerical proxy for the exact clean endpoint, explicitly aligning the student's velocity prediction with the corresponding interval-averaged teacher velocity. From a theoretical perspective, we show that this teacher anchor directly constrains the total endpoint error, complementing local consistency supervision with an explicit action endpoint constraint.

To reduce inter-expert waiting, we introduce \textit{Cross-Expert Wavefront Pipelining} (CEWP), which transforms coarse expert-wise execution into a fine-grained block-wise pipeline. Under conventional expert-wise execution, the Action Expert waits for the Video Expert to construct the complete block-wise KV cache. We first conduct an in-depth dependency analysis, observing that each action block requires the video KV cache only from its corresponding video block, rather than the complete output of the Video Expert. Motivated by this observation, CEWP overlaps the execution of the two experts through block-wise sharing of the video KV cache, synchronizing only immediately before the corresponding action attention consumes it, which significantly reduces unnecessary waiting.


RealtimeWAM achieves impressive results on representative MoT-based WAM architectures, including Fast-WAM~\citep{yuan2026fast} and Faster-WAM~\citep{zhao2026faster}, across diverse downstream benchmarks (\eg, LIBERO~\citep{liu2023libero}, LIBERO-Plus~\citep{fei2025liberoplus}, and RoboTwin 2.0~\citep{chen2025robotwin}). On these benchmarks, it delivers approximately $14\sim25\times$ inference acceleration with an average $<1\%$ accuracy drop compared with the original multi-step models. Notably, RealtimeWAM's per-call inference latency is significantly shorter than the corresponding execution time (\eg, 30 Hz control frequency), enabling real-time action generation. These results demonstrate that RealtimeWAM remains effective across diverse robot configurations, task horizons, and distribution conditions, and provide a practical path toward efficient WAM deployment.

To summarize, our contributions are as follows:

$\blacktriangleright$ We propose RealtimeWAM, a one-step asynchronous WAM that addresses both intra-expert iteration and inter-expert waiting in \textit{any} MoT-based WAM for efficient robot manipulation. \\
$\blacktriangleright$ We propose \textit{Teacher-Anchored Consistency Distillation} (TACD), which mitigates the \textit{local-global error gap} by augmenting local consistency supervision with a teacher endpoint constraint for accurate one-step action generation. \\
$\blacktriangleright$ We introduce \textit{Cross-Expert Wavefront Pipelining} (CEWP), which reduces unnecessary synchronization by overlapping block-wise execution of the Video and Action Experts according to their data dependencies. \\
$\blacktriangleright$ Extensive experiments demonstrate that RealtimeWAM achieves real-time action generation with $<1\%$ accuracy degradation (\ie, 12.2~ms inference latency with $\sim25\times$ speedup on an NVIDIA H100 GPU).

\section{Related Work}

\subsection{World Action Models}
While Vision-Language-Action (VLA) models~\citep{shukor2025smolvla,intelligence2025pi_,kim2024openvla,bjorck2025gr00t,black2024pi_0,liu2025rdt,wang2026vla,zheng2026x} rely on pretrained vision-language backbones to map visual observations and language instructions directly to actions, World Action Models (WAMs)~\citep{yuan2026fast,ye2026world,bi2026motus,li2026causal} incorporate future visual representations from video generation backbones to guide action prediction. Modern WAMs can be broadly classified into two categories. \ding{192} One line of work focuses on shared-backbone WAMs, which model future video and robot actions within a shared backbone, including DreamZero~\citep{ye2026world} and Cosmos~\citep{kim2026cosmos}. \ding{193} In contrast, Mixture-of-Transformers (MoT)-based WAMs, such as Motus~\citep{bi2026motus}, employ separate video and action experts that interact through attention. Among these methods, Fast-WAM~\citep{yuan2026fast} reduces inference overhead by computing video representations \textit{once} for reuse by the action expert. Subsequent works, including Light-WAM~\citep{li2026light} and Faster-WAM~\citep{zhao2026faster}, further improve computational efficiency and out-of-distribution (OOD) robustness. Nevertheless, these methods may incur substantial performance degradation on challenging benchmarks or achieve limited speedups. 

\subsection{Few-Step Distillation}
Diffusion models~\citep{sohl2015deep,ho2020denoising,song2020score} and flow models~\citep{lipman2022flow,liu2022flow,albergo2023stochastic} have become cornerstones of modern high-quality image and video generation. However, their iterative sampling cost has motivated a series of few-step acceleration methods, including progressive distillation~\citep{salimans2022progressive}, consistency models~\citep{song2023consistency}, distribution matching distillation~\citep{yin2024improved}, MeanFlow~\citep{geng2025meanflow}, and adversarial distillation~\citep{sauer2024adversarial}. These distillation techniques have recently been introduced into World Action Models~\citep{akbari2026flash,yang2026wam}. The work most relevant to ours is Flash-WAM~\citep{akbari2026flash}, which identifies a modality-dependent noise-regime mismatch and introduces modality-specific consistency parameterizations. However, \ding{192} it focuses on few-step distillation for WAMs that retain future-video denoising during inference, such as LingBot-VA~\citep{li2026causal}; and \ding{193} its ultra-few-step (\ie, one-step) distillation still exhibits noticeable performance degradation (\eg, RoboTwin and real-world tasks). These limitations motivate our investigation of one-step action distillation for MoT-based WAMs.



\subsection{Asynchronous Inference for Robot Policies}

Asynchronous inference has been widely studied for real-time robot control because it mitigates execution stalls, stale observations, and delayed responses caused by policy latency. Existing approaches can be broadly grouped into three categories: execution-level asynchrony, module-level asynchrony, and sampling-level asynchrony. Execution-level methods overlap prediction of the next action chunk with execution of the current one and maintain continuity through prefix conditioning, state correction, or chunk fusion~\citep{black2025rtc,black2025training,tang2025vlash,wang2026remac,motubrain2026asyncwam}. Module-level methods decouple a slow semantic or world module from a fast action module, allowing cached high-level representations to guide high-frequency action updates~\citep{hirose2026asyncvla,cai2026aha}. Sampling-level methods assign different action positions, chunks, modalities, or denoising streams distinct sampling schedules, enabling early action output or selective refinement without waiting for full decoding~\citep{jiang2025asyncvla,lu2026faster,li2026flashvla}. In contrast, RealtimeWAM employs finer-grained asynchrony at the block level, distinguishing it from prior approaches.

\section{Preliminaries}
\subsection{Efficient Mixture-of-Transformers (MoT)-based WAMs}
\label{sec:fast_wam}
WAMs with a Mixture-of-Transformers (MoT) architecture~\citep{bi2026motus,li2026causal} comprise a pretrained Video Expert for world modeling and an Action Expert for action generation. Corresponding blocks in the two experts interact through shared attention. While the two experts are jointly trained on video and action prediction, recent efficient MoT-based WAMs~\citep{yuan2026fast,zhao2026faster} require only a single forward pass through the Video Expert during inference. Specifically, the Video Expert processes the current observation $o$ and language instruction $l$ to extract block-wise keys and values:
\begin{equation}
    z(o,l) = F_{\psi}^{\mathrm{KV}}(o,l) = \left\{(K_i^v,V_i^v)\right\}_{i=1}^{N}.
    \label{eq:video_kv}
\end{equation}
Here, $F_{\psi}^{\mathrm{KV}}$ denotes the KV extraction function of the Video Expert with parameters $\psi$. The latent world representation $z(o,l)$ consists of keys $K_i^v$ and values $V_i^v$ from each of the $N$ video transformer blocks. Subsequently, these efficient MoT-based WAMs use this representation to parameterize the action distribution:
\begin{equation}
    p_{\theta}(a\mid o,l) = p_{\theta}\!\left(a\mid z(o,l)\right).
    \label{eq:action_distribution}
\end{equation}
Here, $a$ denotes an action chunk with prediction horizon $H$, and $\theta$ denotes the parameters of the Action Expert. Action generation requires $n$ denoising steps along $0=t_0<\cdots<t_n=1$. The action distribution $p_\theta\!\left(a\mid z(o,l)\right)$ is implicitly defined by the denoising map $\Phi_\theta^z=\Phi_{\theta,t_1}^z\circ\Phi_{\theta,t_2}^z\circ\cdots\circ\Phi_{\theta,t_n}^z$, where $a_{t_{j-1}}=\Phi_{\theta,t_j}^z(a_{t_j})=a_{t_j}-(t_j-t_{j-1})v_\theta(a_{t_j},t_j)$. Here, $a_{t_j}$ denotes a noisy action chunk and $v_\theta$ the predicted velocity.

\subsection{Consistency Distillation for Action Generation}
Consistency distillation (CD)~\citep{song2023consistency} learns to map noisy states (\ie, action chunks in our paper) along the same teacher denoising trajectory to a common clean endpoint (\ie, $t=0$), enabling few-step generation. It uses three models: a frozen teacher $\theta_{\mathrm T}$, a trainable student $\theta_{\mathrm S}$, and an exponential moving average (EMA) target $\theta_{\mathrm{ema}}$. During training, given a clean action chunk $a_0^{\mathrm{data}}$ from the training data, we randomly sample $t\in[0,1]$ and Gaussian noise $\epsilon\sim\mathcal{N}(0,I)$ to construct $a_t=(1-t)a_0^{\mathrm{data}}+t\epsilon$. The student and EMA target use the endpoint mapping
\begin{equation}
    f_\theta(a_t,t)=a_t-t\,v_\theta(a_t,t).
    \label{eq:cd_prediction}
\end{equation}
For a neighboring time $0\le s<t$, the teacher produces
\begin{equation}
    \tilde{a}_s = \texttt{Solver}(a_t,t,s;\theta_{\mathrm T}) \approx a_t-(t-s)\,v_{\theta_{\mathrm T}}(a_t,t),
    \label{eq:cd_target}
\end{equation}
where $\texttt{Solver}(a_t,t,s;\theta_{\mathrm T})$ denotes one numerical integration step from $t$ to $s$ using the frozen teacher $\theta_{\mathrm T}$. For neighboring time steps with small $t-s$, treating the velocity as locally constant yields the first-order approximation above. The student is optimized with
\begin{equation}
    \mathcal{L}_{\mathrm{CD}}
    =\mathbb{E}_{a_t,t,s}\!\left[
        \left\|
        f_{\theta_{\mathrm S}}(a_t,t)
        -\operatorname{sg}\!\left[
            f_{\theta_{\mathrm{ema}}}(\tilde{a}_s,s)
        \right]
        \right\|_2^2
    \right],
    \label{eq:cd_loss}
\end{equation}
where $\operatorname{sg}$ denotes stop-gradient. After each student update, $\theta_{\mathrm{ema}}\leftarrow\mu\theta_{\mathrm{ema}}+(1-\mu)\theta_{\mathrm S}$, with decay rate $\mu\in[0,1)$. Therefore, consistency distillation learns to map noisy action chunks directly to clean endpoints, enabling few-step action generation without traversing the full teacher trajectory.

\section{RealtimeWAM}

In this work, we propose \textit{RealtimeWAM}, a one-step asynchronous world action model for post-training acceleration of \textit{any} MoT-based WAM. RealtimeWAM addresses intra-expert iteration through Teacher-Anchored Consistency Distillation (Sec.~\ref{sec:tacd}) and inter-expert waiting through Cross-Expert Wavefront Pipelining (Sec.~\ref{sec:cewp}).

\subsection{Teacher-Anchored Consistency Distillation}
\label{sec:tacd}

\begin{figure}[!t]
    \centering
    \vspace{-0.1in}
    \includegraphics[width=0.9\textwidth]{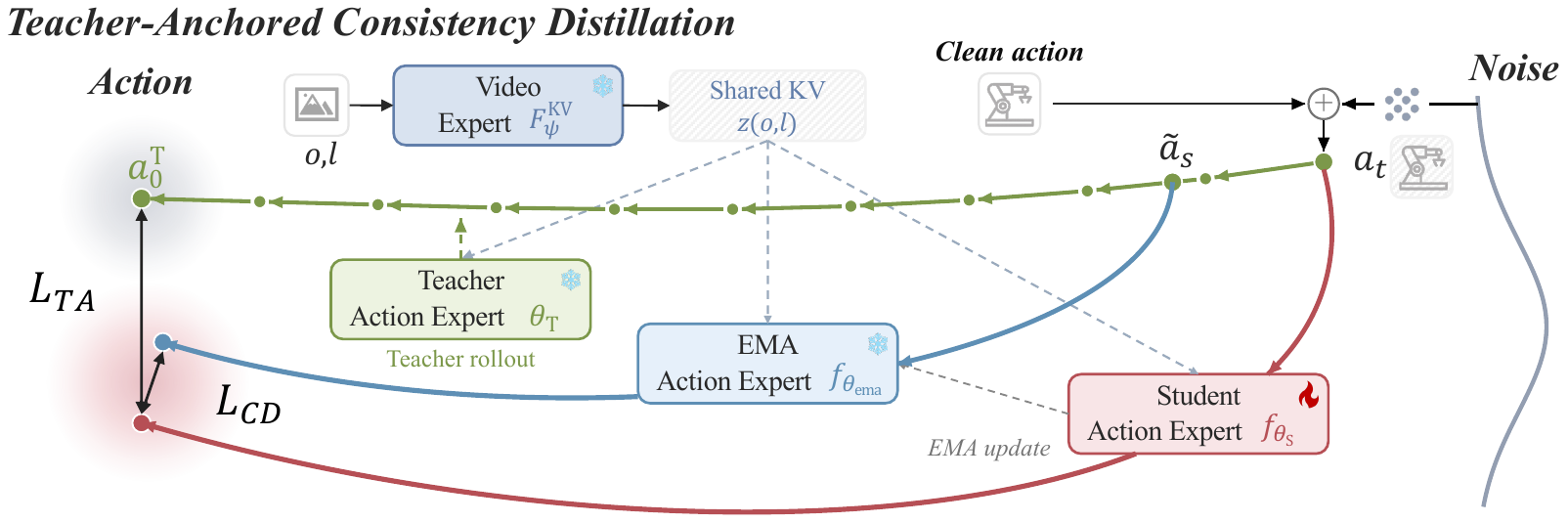}
    \caption{Overview of Teacher-Anchored Consistency Distillation (TACD). The frozen Video Expert provides shared KV conditioning, while the action student is trained with local consistency supervision and a teacher-anchored constraint derived from the frozen teacher's multi-step rollout.}
    \vspace{-0.1in}
    \label{fig:method1}
\end{figure}



As discussed above, multi-step action denoising dominates inference latency (Fig.~\ref{fig:intro}) while video latent representations are computed in a single forward pass in the Video Expert. Therefore, our target is to achieve high-quality few-step action generation (especially one-step) to combat intra-expert iteration. Recent step distillation strategies~\citep{song2023consistency,yin2024improved,geng2025meanflow,huang2026meanflownft} have been extensively studied in image and video generation. However, we identify a distinctive characteristic of action generation in WAMs: denoising is strongly conditioned on visual observations, robot states, and language instructions, which can confine feasible actions (\eg, grasp positions and gripper commands) to a narrow range. Within this constrained action space, even subtle deviations can affect task success rates owing to the high sensitivity of robotic control to action errors. 

\begin{wrapfigure}{r}{0.60\linewidth}
    \vspace{-0.1in}
    \centering
    \setlength{\abovecaptionskip}{0.2cm}
    \includegraphics[width=\linewidth]{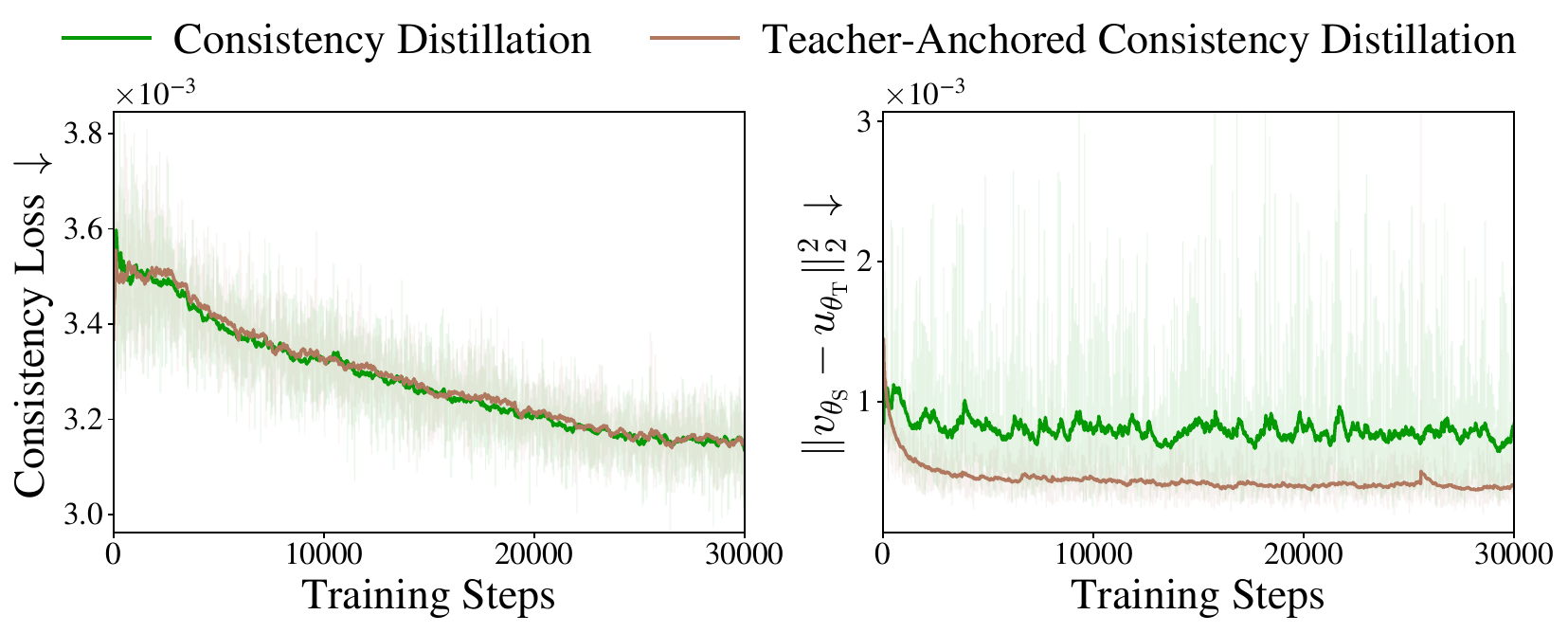}
    \caption{Training curves of two distillation methods: consistency loss (\textit{Left}) and squared distance between the student's velocity $v_{\theta_{\mathrm S}}$ and the teacher's average velocity towards the clean endpoint $u_{\theta_{\mathrm T}}$ (\textit{Right}).}
    \label{fig:loss_curves}
    \vspace{-0.1in}
\end{wrapfigure}
However, directly applying consistency distillation to WAMs raises a significant challenge, dubbed the \textit{local-global error gap}. Specifically, although consistency distillation~\citep{song2023consistency} can effectively minimize \textit{local} consistency errors, it does not considerably reduce deviations from the \textit{global} target action. As shown in Fig.~\ref{fig:loss_curves} (right), the error between the teacher's clean action targets and student's predictions does not decrease considerably for consistency distillation. Such large deviations from the clean action targets can induce inaccuracies in robotic control.

Formally, let $a_t$ denote a noisy action chunk and $\tilde{a}_s$ the teacher-updated action defined analogously to Eq.~\eqref{eq:cd_target}. Let $a_0^\star$ denote the exact clean endpoint of the frozen teacher ODE initialized at $a_t$, where $\star$ indicates the theoretical exact solution. Omitting the shared conditioning $z(o,l)$ for brevity, the \textit{total} error decomposes into a \textit{local} consistency error and a \textit{global} target error:
\begin{equation}
    \underbrace{f_{\theta_{\mathrm S}}(a_t,t)-a_0^\star}_{\mathbf{e}_{\mathrm{total}}:\ \text{Total error}} = \underbrace{f_{\theta_{\mathrm S}}(a_t,t)-f_{\theta_{\mathrm{ema}}}(\tilde{a}_s,s)}_{\mathbf{e}_{\mathrm{local}}:\ \text{Local consistency error}} + \underbrace{f_{\theta_{\mathrm{ema}}}(\tilde{a}_s,s)-a_0^\star}_{\mathbf{e}_{\mathrm{global}}:\ \text{Global target error}}.
    \label{eq:action_error_decomposition}
\end{equation}
Consistency Distillation directly minimizes $\mathbb{E}\|\mathbf{e}_{\mathrm{local}}\|_2^2$ but does not explicitly penalize $\mathbf{e}_{\mathrm{global}}$. Consequently, a small \textit{local} consistency error alone does not guarantee a small total error $\mathbf{e}_{\mathrm{total}}$ (see Appendix~\ref{app:local_global_error}). Therefore, our key insight is to preserve \textit{local} consistency supervision while introducing an additional constraint based on the teacher's target action (\ie, $a_0^\star$), thereby mitigating the \textit{global} deviation that the local objective may leave unresolved.

Motivated by this insight, we propose \textit{Teacher-Anchored Consistency Distillation} (TACD), a novel post-training step distillation framework tailored for MoT-based WAMs. Specifically, since the exact endpoint $a_0^\star$ is unavailable in practice, we adopt the frozen teacher's multi-step rollout endpoint $a_0^{\mathrm T}$ as a proxy and compute the corresponding interval-averaged velocity. It is worth noting that the three Action Experts (\ie, the frozen teacher, trainable student, and EMA target) share the same video KV cache $z(o,l)$ from a single frozen Video Expert. Starting from the same noisy action $a_t$ under this shared conditioning $z(o,l)$, we have
\begin{equation}
    a_0^{\mathrm T} = \texttt{Solver}(a_t,t,0;\theta_{\mathrm T}), \qquad u_{\theta_{\mathrm T}}(a_t,t) = \frac{a_t-a_0^{\mathrm T}}{t}, \quad t>0,
    \label{eq:teacher_rollout}
\end{equation}
where $\texttt{Solver}$ denotes numerical integration from $t$ to $0$ using $K$ teacher denoising steps.

We further directly supervise the student's velocity prediction $v_{\theta_{\mathrm S}}(a_t,t)$ with the teacher's interval-averaged velocity $u_{\theta_{\mathrm T}}(a_t,t)$, yielding the teacher-anchored loss and the overall training objective:
\begin{equation}
    \mathcal{L}_{\mathrm{TA}} = \mathbb{E}_{a_t,t}\!\left[\left\|v_{\theta_{\mathrm S}}(a_t,t)-\operatorname{sg}\!\left[u_{\theta_{\mathrm T}}(a_t,t)\right]\right\|_2^2\right], \qquad \mathcal{L} = \mathcal{L}_{\mathrm{CD}}+\lambda\mathcal{L}_{\mathrm{TA}},
    \label{eq:tacd_loss}
\end{equation}
where $\lambda>0$ weights the teacher-anchored loss. Equivalently, $\mathcal{L}_{\mathrm{TA}}$ directly supervises the student's clean action prediction against the teacher's rollout endpoint, with weight $1/t^2$ (see Appendix~\ref{app:weighted_endpoint_supervision}).

To relate this supervision to the exact teacher endpoint, let $\boldsymbol{\delta}=a_0^{\mathrm T}-a_0^\star$ denote the teacher's numerical integration error. The total endpoint error in Eq.~\eqref{eq:action_error_decomposition} then satisfies
\begin{equation}
    \mathbf{e}_{\mathrm{total}} = \mathbf{e}_{\mathrm{local}}+\mathbf{e}_{\mathrm{global}} = -t\!\left[v_{\theta_{\mathrm S}}(a_t,t)-u_{\theta_{\mathrm T}}(a_t,t)\right]+\boldsymbol{\delta}.
    \label{eq:ta_total_error}
\end{equation}
Consequently, teacher anchoring directly constrains the combined error $\mathbf{e}_{\mathrm{local}}+\mathbf{e}_{\mathrm{global}}$, complementing the local consistency objective with a direct endpoint constraint\footnote{The total endpoint error $\mathbf{e}_{\mathrm{total}}$ also depends on the teacher's numerical integration error $\boldsymbol{\delta}$, which can be controlled by increasing the number of teacher denoising steps $K$ (see Appendix~\ref{app:teacher_anchoring_bound}).}. In addition, for $0<t\le1$, this relation yields an upper bound on the expected squared endpoint error, $\mathbb{E}\|\mathbf{e}_{\mathrm{total}}\|_2^2 \le 2\mathcal{L}_{\mathrm{TA}}+2\mathbb{E}\|\boldsymbol{\delta}\|_2^2$ (see Appendix~\ref{app:teacher_anchoring_bound}).


\subsection{Cross-Expert Wavefront Pipelining}
\label{sec:cewp}

\begin{figure}[!t]
    \centering
    \vspace{-0.1in}
    \includegraphics[width=1\textwidth]{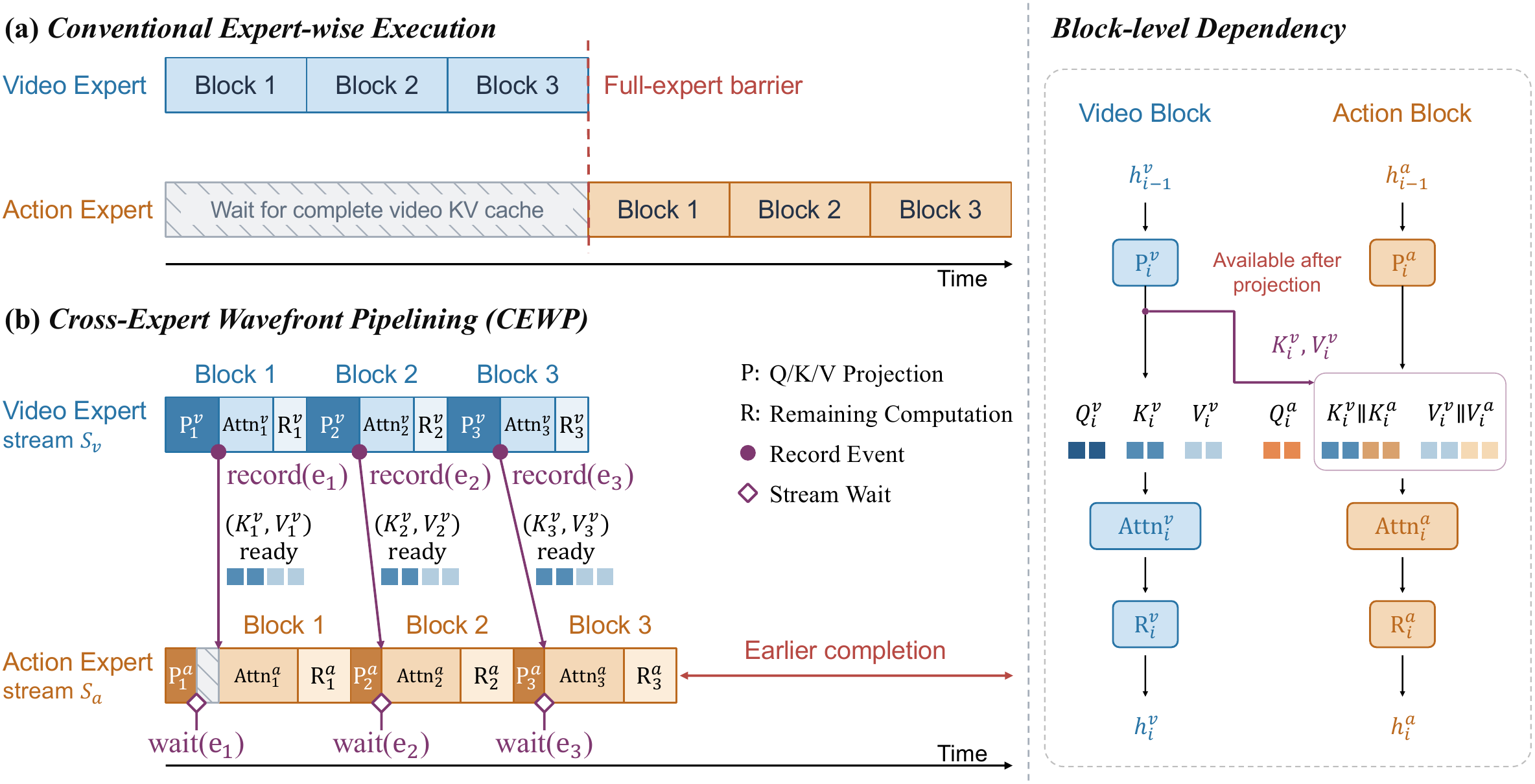}
    \caption{Comparison of expert execution schedules. (a) Conventional execution waits for the complete video KV cache. (b) Cross-Expert Wavefront Pipelining releases block-wise KV cache entries after projection and synchronizes only before action attention consumes them, overlapping the Video and Action Experts.}
    \label{fig:method2}
    \vspace{-0.1in}
\end{figure}

Teacher-Anchored Consistency Distillation reduces intra-expert iteration by enabling one-step action generation. However, inter-expert waiting persists because the Video and Action Experts still execute sequentially (Fig.~\ref{fig:method2}(a)). Specifically, the Action Expert waits for the complete video KV cache, creating a global synchronization barrier that prevents action computation from starting even when its required context is already available.

We therefore analyze the actual data dependencies between the Video Expert and Action Expert in a fine-grained manner. We first decompose each DiT block into three sequential components: Q/K/V projection $\mathcal{P}_i$, attention $\operatorname{Attn}_i$, and the remaining computation $\mathcal{R}_i$. Let $h_{i-1}^v$ and $h_{i-1}^a$ denote the video and action hidden states entering block $i\in\{1,\ldots,N\}$, respectively. Their projections are
\begin{equation}
    (Q_i^v,K_i^v,V_i^v)=\mathcal{P}_i^v(h_{i-1}^v), \qquad (Q_i^a,K_i^a,V_i^a)=\mathcal{P}_i^a(h_{i-1}^a).
    \label{eq:expert_projections}
\end{equation}
Here, $(K_i^v,V_i^v)$ constitute the $i$-th entry of $z(o,l)$ in Eq.~\eqref{eq:video_kv}. The Video Expert attends to its own tokens, while the Action Expert additionally incorporates the corresponding video KV cache:
\begin{equation}
    \begin{aligned}
        h_i^v &= \mathcal{R}_i^v\!\left(h_{i-1}^v,\operatorname{Attn}_i^v\!\left(Q_i^v,K_i^v,V_i^v\right)\right), \\
        h_i^a &= \mathcal{R}_i^a\!\left(h_{i-1}^a,\operatorname{Attn}_i^a\!\left(Q_i^a,K_i^v\mathbin{\Vert}K_i^a,V_i^v\mathbin{\Vert}V_i^a\right)\right),
    \end{aligned}
    \label{eq:expert_block_updates}
\end{equation}
where $\mathbin{\Vert}$ denotes token-wise concatenation. Action attention $\operatorname{Attn}_i^a$ is thus the only operation in action block $i$ that directly consumes the video KV cache $(K_i^v,V_i^v)$, with the dependency
\begin{equation}
    \mathcal{P}_i^v \longrightarrow \operatorname{Attn}_i^a \longleftarrow \mathcal{P}_i^a.
    \label{eq:cross_expert_dependency}
\end{equation}
Our key observations are: \ding{192} The action projection $\mathcal{P}_i^a$ can execute once $h_{i-1}^a$ is available, without waiting for the video KV cache $(K_i^v,V_i^v)$. \ding{193} Action attention $\operatorname{Attn}_i^a$ can begin once both $\mathcal{P}_i^a$ and $\mathcal{P}_i^v$ complete, without waiting for video attention $\operatorname{Attn}_i^v$ or video states $h_j^v$ with $j\geq i$. \ding{194} Existing methods~\citep{yuan2026fast,zhao2026faster} enforce $\mathcal{R}_N^v\rightarrow\mathcal{P}_1^a$, introducing unnecessary waiting.

Based on this dependency analysis, we propose \textit{Cross-Expert Wavefront Pipelining} (CEWP), which transforms coarse expert-wise execution into a fine-grained block-wise wavefront across the Video and Action Experts. We assign the Video and Action Experts to two non-blocking CUDA streams, $\mathcal{S}_v$ and $\mathcal{S}_a$, respectively. Each stream preserves program order, while a CUDA event $e_i$ enforces the dependency at block $i$:
\begin{equation}
    \begin{aligned}
        \mathcal{S}_v:\;& \mathcal{P}_i^v \rightarrow \operatorname{record}(e_i) \rightarrow \operatorname{Attn}_i^v \rightarrow \mathcal{R}_i^v, \\
        \mathcal{S}_a:\;& \mathcal{P}_i^a \rightarrow \operatorname{wait}(e_i) \rightarrow \operatorname{Attn}_i^a \rightarrow \mathcal{R}_i^a,
    \end{aligned}
    \label{eq:wavefront_schedule}
\end{equation}
where $e_i$ signals the availability of $(K_i^v,V_i^v)$. The video stream records $e_i$ immediately after $\mathcal{P}_i^v$ and continues its remaining computation, while the action stream waits only before $\operatorname{Attn}_i^a$ (shown in Fig.~\ref{fig:method2}(b)). This late synchronization allows the action projection to proceed independently and delays cross-stream waiting until the video KV cache is actually consumed. Action blocks thus consume video KV cache entries incrementally as the video stream advances, forming a staggered wavefront along model depth.

\section{Experiments}
\label{sec:experiments}

\subsection{Implementation Details}
\label{lab:setup}

We implement RealtimeWAM based on Fast-WAM~\citep{yuan2026fast} and Faster-WAM~\citep{zhao2026faster}, yielding RealtimeWAM\textsuperscript{*} and RealtimeWAM\textsuperscript{\textdagger}, respectively. We freeze the Video Expert and optimize only rank-128 low-rank adaptation (LoRA)~\citep{hu2021lora} parameters in the Action Expert. In Teacher-Anchored Consistency Distillation (TACD), we obtain the teacher endpoint through a 10-step rollout of the frozen teacher and set the teacher-anchored loss weight to $\lambda=0.2$. Training proceeds for 30{,}000 iterations with AdamW, a learning rate of $10^{-4}$, and a weight decay of $10^{-2}$. We set the exponential moving average (EMA) decay to $0.995$. For a fair comparison, all evaluated step-distillation methods, including distribution matching distillation (DMD)~\citep{yin2024improved}, consistency distillation~\citep{song2023consistency}, and MeanFlow distillation~\citep{geng2025meanflow,gu2026anyflow}, adopt the same training settings. Moreover, we compare against other efficient WAMs, including Flash-WAM~\citep{akbari2026flash} and Light-WAM~\citep{li2026light}. For latency evaluation, we include VAE encoding time but exclude text encoding time, as instructions are encoded only once per episode.

\subsection{Evaluation Benchmarks}

We evaluate RealtimeWAM on LIBERO~\citep{liu2023libero}, RoboTwin 2.0~\citep{chen2025robotwin}, and LIBERO-Plus~\citep{fei2025liberoplus}, using task success rate as the evaluation metric. On LIBERO, we evaluate four task suites: Spatial, Object, Goal, and Long. For RoboTwin 2.0, we follow the multi-task training setting in previous work~\citep{yuan2026fast}, training on 50 tasks using a total of 2{,}500 clean demonstrations and 25{,}000 randomized demonstrations. We evaluate performance under both clean and randomized conditions. To assess out-of-distribution (OOD) robustness, we further evaluate success rates on all seven perturbation subsets of LIBERO-Plus. We additionally evaluate RealtimeWAM on real-world dual-arm manipulation tasks (see Appendix~\ref{app:garment_folding}).

\subsection{Main Results}

\textbf{RoboTwin 2.0.} We report detailed results on RoboTwin 2.0 (Tab.~\ref{tab:main_results}). Notably, RealtimeWAM\textsuperscript{*} achieves a success rate of 90.84\% with one-step action generation using only LoRA fine-tuning, incurring a reduction of just 0.67\% compared with 10-step Fast-WAM. Surprisingly, RealtimeWAM\textsuperscript{\textdagger} achieves nearly lossless performance with one-step action generation, attaining a success rate of 92.64\% with a reduction of only 0.29\% compared with 10-step Faster-WAM. It also considerably outperforms other efficient WAMs (\eg, 81.41\% for one-step Flash-WAM) and step-distillation baselines (\eg, 88.22\% for DMD\textsuperscript{*}).

\textbf{LIBERO.} A similar trend is observed on LIBERO (Tab.~\ref{tab:main_results}). RealtimeWAM\textsuperscript{*} achieves an overall success rate of 97.0\% with one-step action generation, matching the performance of 10-step Fast-WAM. Compared with directly reducing Fast-WAM to a single denoising step (95.0\%), RealtimeWAM\textsuperscript{*} improves the success rate by 2.0\%, demonstrating the effectiveness of step distillation in preserving policy performance. It also outperforms one-step Flash-WAM (95.1\%) and DMD\textsuperscript{*} (96.1\%), while remaining competitive with Light-WAM (97.2\%). Detailed results for individual LIBERO task suites are provided in Appendix~\ref{app:libero_results} (Tab.~\ref{tab:libero}).

\begin{table*}[t]
    \centering
    \setlength{\tabcolsep}{6pt}
    \renewcommand{\arraystretch}{1.05}
    \caption{Success rates (\%) on RoboTwin 2.0 and LIBERO. NFE denotes the number of denoising steps for video and action generation. Methods marked with \textsuperscript{*} and \textsuperscript{\textdagger} are built on Fast-WAM and Faster-WAM, respectively.}
    \label{tab:main_results}
    \resizebox{\linewidth}{!}{
\begin{tabular}[t]{l|cc|ccc|c}
\toprule
\multirow{2}{*}{Method}
& \multicolumn{2}{c|}{NFE}
& \multicolumn{3}{c|}{RoboTwin 2.0} & LIBERO \\
& Video & Action
& Clean$\uparrow$ & Random$\uparrow$ & Overall$\uparrow$ & Overall$\uparrow$ \\
\midrule
\multicolumn{7}{c}{\cellcolor[gray]{0.92}WAM and VLA Baselines} \\
\midrule
$\pi_{0}$~\citep{black2024pi_0}       & -- & -- & 65.92 & 58.40 & 62.16 & 94.1 \\
$\pi_{0.5}$~\citep{intelligence2025pi_}     & -- & -- & 82.74 & 76.76 & 79.75 & 96.9 \\
X-VLA~\citep{zheng2026x}           & -- & -- & 72.90 & 72.80 & 72.85 & 98.1 \\
Motus~\citep{bi2026motus}           & 10 & 10 & 88.66 & 87.02 & 87.84 & 97.7 \\
LingBot-VA~\citep{li2026causal}      & 25 & 50 & 92.90 & 91.50 & 92.20 & 98.5 \\
Fast-WAM~\citep{yuan2026fast}         & 1 & 10 & 91.82 & 91.19 & 91.51 & 97.0 \\
Faster-WAM~\citep{zhao2026faster}    & 1 & 10 & 93.20 & 92.66 & 92.93 & 98.9 \\
\midrule
\multicolumn{7}{c}{\cellcolor[gray]{0.92}Other Efficient Variants} \\
\midrule
Flash-WAM~\citep{akbari2026flash}       & 1 & 2 & 88.42 & 82.66 & 85.54 & 96.1 \\
Flash-WAM~\citep{akbari2026flash}       & 1 & 1 & 82.56 & 80.26 & 81.41 & 95.1 \\
Light-WAM~\citep{li2026light}       & 1 & 1 & 76.40 & 76.30 & 76.40 & 97.2 \\
\midrule
\multicolumn{7}{c}{\cellcolor[gray]{0.92}Few-Step Distillation} \\
\midrule
Fast-WAM~\citep{yuan2026fast}         & 1 & 1  & 70.42 & 68.12 & 69.27 & 95.0 \\
Mean Flow\textsuperscript{*}~\citep{geng2025meanflow} & 1 & 1 & 85.74 & 85.72 & 85.73 & 96.9 \\
Consistency Distillation\textsuperscript{*}~\citep{song2023consistency} & 1 & 1 & 89.88 & 89.38 & 89.63 & 96.9 \\
DMD\textsuperscript{*}~\citep{yin2024improved} & 1 & 1 & 88.64 & 87.80 & 88.22 & 96.1 \\
\rowcolor{mycolor!30}
RealtimeWAM\textsuperscript{*}  & 1 & 1 & 91.96 & 89.72 & 90.84 & 97.0 \\
\rowcolor{mycolor!30}
RealtimeWAM\textsuperscript{\textdagger} & 1 & 1 & 92.98 & 92.30 & 92.64 & 99.0 \\
\bottomrule
\end{tabular}

    }
    \vspace{-0.1in}
\end{table*}

\textbf{LIBERO-Plus.} We further evaluate robustness to distribution shifts on LIBERO-Plus (Tab.~\ref{tab:libero_plus}). With one-step action generation, RealtimeWAM\textsuperscript{\textdagger} achieves an overall success rate of 73.0\%, only 0.6\% below Faster-WAM (73.6\%). It slightly improves performance under camera, sensor-noise, and layout perturbations, indicating that step distillation largely preserves out-of-distribution robustness.

\begin{table*}[t]
    \centering
    \setlength{\tabcolsep}{3pt}
    \renewcommand{\arraystretch}{1.05}
    \caption{Success rates (\%) on the seven perturbation subsets of LIBERO-Plus.}
    \label{tab:libero_plus}
    \resizebox{\linewidth}{!}{
\begin{tabular}[t]{l|ccccccc|c}
\toprule
Method & Camera$\uparrow$ & Robot$\uparrow$ & Lang.$\uparrow$ & Light$\uparrow$
& Backg.$\uparrow$ & Noise$\uparrow$ & Layout$\uparrow$ & Overall$\uparrow$ \\
\midrule
UniVLA~\citep{bu2025univla} & 1.8 & 46.2 & 69.6 & 69.0 & 81.0 & 21.2 & 31.9 & 42.9 \\
OpenVLA-OFT~\citep{kim2025fine} & 56.4 & 31.9 & 79.5 & 88.7 & 93.3 & 75.8 & 74.2 & 69.6 \\
$\pi_{0}$~\citep{black2024pi_0} & 13.8 & 6.0 & 58.8 & 85.0 & 81.4 & 79.0 & 68.9 & 53.6 \\
$\pi_{0}$-Fast~\citep{pertsch2025fast} & 65.1 & 21.6 & 61.0 & 73.2 & 73.2 & 74.4 & 68.8 & 61.6 \\
WorldVLA~\citep{cen2025worldvla} & 0.1 & 27.9 & 41.6 & 43.7 & 17.1 & 10.9 & 38.0 & 25.0 \\
Fast-WAM~\citep{yuan2026fast} & 18.8 & 45.7 & 70.1 & 83.2 & 45.7 & 29.8 & 62.7 & 49.1 \\
Faster-WAM~\citep{zhao2026faster} & 53.8 & 71.6 & 94.7 & 96.3 & 61.3 & 63.6 & 79.1 & 73.6 \\
\midrule
\rowcolor{mycolor!30}
RealtimeWAM\textsuperscript{\textdagger} & 54.2 & 69.0 & 92.5 & 94.7 & 59.4 & 64.6 & 79.3 & 73.0 \\
\bottomrule
\end{tabular}

    }
    \vspace{-0.1in}
\end{table*}

\subsection{Ablation Study}
\label{lab:ablaion}
\begin{table}[!ht]
    \centering
    \begin{minipage}[b]{0.48\linewidth}
        \centering
        \setlength{\tabcolsep}{3pt}
        \renewcommand{\arraystretch}{1.1}
        \caption{Ablation of loss terms and expert fine-tuning on RoboTwin 2.0. $\mathcal{L}_{\mathrm{CD}}$ and $\mathcal{L}_{\mathrm{TA}}$ denote the consistency and teacher-anchored losses, respectively. Video/Action indicate whether each expert is fine-tuned. Results are success rates (\%); best results are in bold.}
        \label{tab:ablation_study}
        \resizebox{\linewidth}{!}{
\begin{tabular}{ccccccc}
\toprule
$\mathcal{L}_{\mathrm{CD}}$ & $\mathcal{L}_{\mathrm{TA}}$
& Video & Action & Clean$\uparrow$ & Random$\uparrow$ & Overall$\uparrow$ \\
\midrule
\ding{51} & \ding{55} & \ding{51} & \ding{51} & 89.88 & 89.38 & 89.63 \\
\ding{51} & \ding{55} & \ding{55} & \ding{51} & 89.76 & \textbf{89.94} & 89.85 \\
\rowcolor{mycolor!30}
\ding{51} & \ding{51} & \ding{55} & \ding{51} & \textbf{91.96} & 89.72 & \textbf{90.84} \\
\bottomrule
\end{tabular}
}
    \end{minipage}\hfill
    \begin{minipage}[b]{0.48\linewidth}
        \centering
        \setlength{\tabcolsep}{12pt}
        \renewcommand{\arraystretch}{1.1}
        \caption{Effect of teacher rollout steps $K$ in TACD on RoboTwin 2.0. $K$ denotes the number of teacher denoising steps used to obtain $a_0^{\mathrm T}$. We use $K=10$ by default. Results are success rates (\%); best results are in bold.}
        \label{tab:teacher_steps}
        \resizebox{\linewidth}{!}{\begin{tabular}{cccc}
\toprule
$K$ & Clean$\uparrow$ & Random$\uparrow$ & Overall$\uparrow$ \\
\midrule
5 & 91.02 & 89.76 & 90.39 \\
\rowcolor{mycolor!30}
10 & \textbf{91.96} & 89.72 & \textbf{90.84} \\
20 & 91.22 & \textbf{90.32} & 90.77 \\
\bottomrule
\end{tabular}
}
    \end{minipage}
\end{table}

\textbf{Effect of different components.} We first validate the effectiveness of the key components of our distillation framework (Tab.~\ref{tab:ablation_study}). Fine-tuning the Video Expert during action step distillation reduces the overall success rate while introducing additional training overhead (\ie, longer training time and higher peak memory usage, as reported in Appendix~\ref{app:limitation}). With the Video Expert frozen, augmenting $\mathcal{L}_{\mathrm{CD}}$ with $\mathcal{L}_{\mathrm{TA}}$, yielding \textit{Teacher-Anchored Consistency Distillation} (TACD), further improves the reported overall success rate from 89.85\% to 90.84\%. This improvement is consistent with our analysis: the teacher anchor provides an endpoint reference from the complete teacher trajectory, supplementing \textit{local} consistency supervision with an explicit endpoint constraint to mitigate \textit{global} deviations that the local objective may leave unresolved.

\textbf{Choice of teacher rollout steps $K$.} We further present results for different numbers of teacher rollout steps $K$ (Tab.~\ref{tab:teacher_steps}). With $K=5$, the rollout may produce a less accurate approximation $a_0^{\mathrm T}$ to the exact clean endpoint $a_0^\star$, resulting in a larger numerical integration error $\boldsymbol{\delta}$ and weaker endpoint supervision. This setting achieves a success rate of 90.39\%, which improves to 90.84\% when $K$ increases to 10. Further increasing $K$ to 20 yields no additional improvement (90.77\%) while requiring more teacher computation. We therefore adopt $K=10$ to balance policy performance and teacher rollout cost.

\begin{figure}[!t]
    \centering
    \vspace{-0.1in}
    \includegraphics[width=\linewidth]{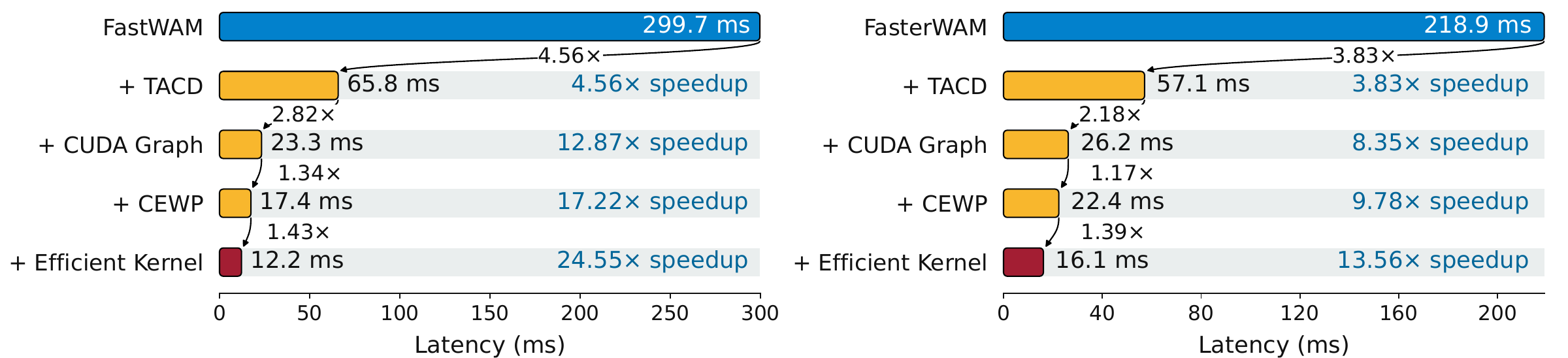}
    \caption{Inference latency of RealtimeWAM built on Fast-WAM (\textit{left}) and Faster-WAM (\textit{right}) on a single NVIDIA H100 GPU.}
    \vspace{-0.1in}
    \label{fig:latency_speedup}
\end{figure}

\textbf{Efficiency Results.} RealtimeWAM reduces inference latency through algorithm and system co-optimization (Fig.~\ref{fig:latency_speedup}). In addition to TACD and CEWP, we employ CUDA Graph and efficient kernels from LightX2V~\citep{lightx2v} (see Appendix~\ref{app:efficient_kernels}). On Fast-WAM and Faster-WAM, TACD reduces latency from 299.7 and 218.9~ms to 65.8 and 57.1~ms, yielding $4.56\times$ and $3.83\times$ speedups, respectively. After enabling CUDA Graph, CEWP further reduces latency from 23.3 to 17.4~ms on Fast-WAM and from 26.2 to 22.4~ms on Faster-WAM, providing additional speedups of $1.34\times$ and $1.17\times$. With all optimizations enabled, RealtimeWAM achieves inference latencies of 12.2 and 16.1~ms, corresponding to overall speedups of $24.55\times$ and $13.56\times$ over the respective baselines. Detailed latency results on other GPUs (\eg, RTX 4090D and RTX 5090) are provided in Appendix~\ref{app:gpu_latency} (Tab.~\ref{tab:gpu_latency}).

\section{Conclusion}
\label{sec:conclusion}

We presented RealtimeWAM, a post-training framework that accelerates MoT-based WAMs by addressing intra-expert iteration and inter-expert waiting. \textit{Teacher-Anchored Consistency Distillation} complements local consistency supervision with a teacher endpoint constraint, enabling accurate one-step action generation. \textit{Cross-Expert Wavefront Pipelining} overlaps block-wise execution of the Video and Action Experts through fine-grained synchronization, reducing unnecessary waiting. RealtimeWAM enables real-time action generation (\eg, 12.2~ms latency) with near-lossless policy performance on MoT-based WAMs, including Fast-WAM and Faster-WAM, while delivering significant end-to-end speedups across diverse hardwares (\eg, approximately $14\sim25\times$ on H100).

\subsection*{AI use statement}
In this work, we used generative AI tools solely to improve the language and readability of the manuscript. We did not use these tools for any tasks requiring disclosure under the ICLR 2027 AI Policy for Authors. The authors reviewed all AI-assisted edits to ensure that they preserved the original technical meaning. We take responsibility for the final content of this work, including the AI-assisted language edits.

\bibliography{iclr2027_conference}

@article{liu2023libero,
  title={{LIBERO}: Benchmarking Knowledge Transfer for Lifelong Robot Learning},
  author={Liu, Bo and Zhu, Yifeng and Gao, Chongkai and Feng, Yihao and Liu, Qiang and Zhu, Yuke and Stone, Peter},
  journal={arXiv preprint arXiv:2306.03310},
  year={2023}
}

@article{chen2025robotwin,
  title={{RoboTwin 2.0}: A Scalable Data Generator and Benchmark with Strong Domain Randomization for Robust Bimanual Robotic Manipulation},
  author={Chen, Tianxing and Chen, Zanxin and Chen, Baijun and Cai, Zijian and Liu, Yibin and Li, Zixuan and Liang, Qiwei and Lin, Xianliang and Ge, Yiheng and Gu, Zhenyu and others},
  journal={arXiv preprint arXiv:2506.18088},
  year={2025}
}

@article{fei2025liberoplus,
  title={{LIBERO-Plus}: In-depth Robustness Analysis of Vision-Language-Action Models},
  author={Fei, Senyu and Wang, Siyin and Shi, Junhao and Dai, Zihao and Cai, Jikun and Qian, Pengfang and Ji, Li and He, Xinzhe and Zhang, Shiduo and Fei, Zhaoye and Fu, Jinlan and Gong, Jingjing and Qiu, Xipeng},
  journal={arXiv preprint arXiv:2510.13626},
  year={2025}
}

@article{yuan2026fast,
  title={Fast-wam: Do world action models need test-time future imagination?},
  author={Yuan, Tianyuan and Dong, Zibin and Liu, Yicheng and Zhao, Hang},
  journal={arXiv preprint arXiv:2603.16666},
  year={2026}
}

@article{ye2026world,
  title={World action models are zero-shot policies},
  author={Ye, Seonghyeon and Ge, Yunhao and Zheng, Kaiyuan and Gao, Shenyuan and Yu, Sihyun and Kurian, George and Indupuru, Suneel and Tan, You Liang and Zhu, Chuning and Xiang, Jiannan and others},
  journal={arXiv preprint arXiv:2602.15922},
  year={2026}
}

@article{shukor2025smolvla,
  title={Smolvla: A vision-language-action model for affordable and efficient robotics},
  author={Shukor, Mustafa and Aubakirova, Dana and Capuano, Francesco and Kooijmans, Pepijn and Palma, Steven and Zouitine, Adil and Aractingi, Michel and Pascal, Caroline and Russi, Martino and Marafioti, Andres and others},
  journal={arXiv preprint arXiv:2506.01844},
  year={2025}
}

@article{intelligence2025pi_,
  title={{$\pi_{0.5}$}: a Vision-Language-Action Model with Open-World Generalization},
  author={Intelligence, Physical and Black, Kevin and Brown, Noah and Darpinian, James and Dhabalia, Karan and Driess, Danny and Esmail, Adnan and Equi, Michael and Finn, Chelsea and Fusai, Niccolo and others},
  journal={arXiv preprint arXiv:2504.16054},
  year={2025}
}

@article{kim2024openvla,
  title={Openvla: An open-source vision-language-action model},
  author={Kim, Moo Jin and Pertsch, Karl and Karamcheti, Siddharth and Xiao, Ted and Balakrishna, Ashwin and Nair, Suraj and Rafailov, Rafael and Foster, Ethan and Lam, Grace and Sanketi, Pannag and others},
  journal={arXiv preprint arXiv:2406.09246},
  year={2024}
}

@article{bjorck2025gr00t,
  title={Gr00t n1: An open foundation model for generalist humanoid robots},
  author={Bjorck, Johan and Casta{\~n}eda, Fernando and Cherniadev, Nikita and Da, Xingye and Ding, Runyu and Fan, Linxi and Fang, Yu and Fox, Dieter and Hu, Fengyuan and Huang, Spencer and others},
  journal={arXiv preprint arXiv:2503.14734},
  year={2025}
}

@article{black2024pi_0,
  title={{$\pi_0$}: A Vision-Language-Action Flow Model for General Robot Control},
  author={Black, Kevin and Brown, Noah and Driess, Danny and Esmail, Adnan and Equi, Michael and Finn, Chelsea and Fusai, Niccolo and Groom, Lachy and Hausman, Karol and Ichter, Brian and others},
  journal={arXiv preprint arXiv:2410.24164},
  year={2024}
}

@inproceedings{liu2025rdt,
  title={Rdt-1b: a diffusion foundation model for bimanual manipulation},
  author={Liu, Songming and Wu, Lingxuan and Li, Bangguo and Tan, Hengkai and Chen, Huayu and Wang, Zhengyi and Xu, Ke and Su, Hang and Zhu, Jun},
  booktitle={International Conference on Learning Representations},
  volume={2025},
  pages={29982--30009},
  year={2025}
}

@inproceedings{bi2026motus,
  title={Motus: A unified latent action world model},
  author={Bi, Hongzhe and Tan, Hengkai and Xie, Shenghao and Wang, Zeyuan and Huang, Shuhe and Liu, Haitian and Zhao, Ruowen and Feng, Yao and Xiang, Chendong and Rong, Yinze and others},
  booktitle={Proceedings of the IEEE/CVF Conference on Computer Vision and Pattern Recognition},
  pages={35101--35113},
  year={2026}
}

@article{li2026causal,
  title={Causal world modeling for robot control},
  author={Li, Lin and Zhang, Qihang and Luo, Yiming and Yang, Shuai and Wang, Ruilin and Han, Fei and Yu, Mingrui and Gao, Zelin and Xue, Nan and Zhu, Xing and others},
  journal={arXiv preprint arXiv:2601.21998},
  year={2026}
}

@article{kim2026cosmos,
  title={Cosmos policy: Fine-tuning video models for visuomotor control and planning},
  author={Kim, Moo Jin and Gao, Yihuai and Lin, Tsung-Yi and Lin, Yen-Chen and Ge, Yunhao and Lam, Grace and Liang, Percy and Song, Shuran and Liu, Ming-Yu and Finn, Chelsea and others},
  journal={arXiv preprint arXiv:2601.16163},
  year={2026}
}

@article{zhao2026faster,
  title={Faster-WAM: Efficient Inference-Time Future Conditioning for Robust World Action Models},
  author={Zhao, Weiheng and Jiang, Haoyi and Shi, Xin and Liu, Liu and Huang, Fan and Su, Zhizhong and Sui, Wei and Wang, Xinggang},
  journal={arXiv preprint arXiv:2608.04404},
  year={2026}
}

@article{li2026light,
  title={Light-WAM: Efficient World Action Models with State-Fusion Action Decoding},
  author={Li, Ziang and Cheng, Dongzhou and Wang, Yibin and Wang, Shiyue and Xu, Xiaoyang and Weng, Lingxuan and Wang, Juan and Wang, Jiaqi},
  journal={arXiv preprint arXiv:2606.08242},
  year={2026}
}

@inproceedings{wang2026vla,
  title={Vla-adapter: An effective paradigm for tiny-scale vision-language-action model},
  author={Wang, Yihao and Ding, Pengxiang and Li, Lingxiao and Cui, Can and Ge, Zirui and Tong, Xinyang and Song, Wenxuan and Zhao, Han and Zhao, Wei and Hou, Pengxu and others},
  booktitle={Proceedings of the AAAI conference on artificial intelligence},
  volume={40},
  number={22},
  pages={18638--18646},
  year={2026}
}

@inproceedings{zheng2026x,
  title={X-vla: Soft-prompted transformer as scalable cross-embodiment vision-language-action model},
  author={Zheng, Jinliang and Li, Jianxiong and Wang, Zhihao and Liu, Dongxiu and Kang, Xirui and Feng, Yuchun and Zheng, Yinan and Zou, Jiayin and Chen, Yilun and Zeng, Jia and others},
  booktitle={International Conference on Learning Representations},
  volume={2026},
  pages={60580--60606},
  year={2026}
}

@article{akbari2026flash,
  title={Flash-WAM: Modality-Aware Distillation for World Action Models},
  author={Akbari, Arman and Zhang, Ci and Akbari, Arash and Zhao, Lin and Chen, Yixiao and Chen, Weiwei and Zhang, Xuan and Yuan, Geng and Wang, Yanzhi},
  journal={arXiv preprint arXiv:2606.05254},
  year={2026}
}

@article{geng2025meanflow,
  title={Mean Flows for One-step Generative Modeling},
  author={Geng, Zhengyang and Deng, Mingyang and Bai, Xingjian and Kolter, J. Zico and He, Kaiming},
  journal={arXiv preprint arXiv:2505.13447},
  year={2025}
}

@inproceedings{song2023consistency,
  title={Consistency Models},
  author={Song, Yang and Dhariwal, Prafulla and Chen, Mark and Sutskever, Ilya},
  booktitle={Proceedings of the 40th International Conference on Machine Learning},
  pages={32211--32252},
  year={2023}
}

@article{yin2024improved,
  title={Improved Distribution Matching Distillation for Fast Image Synthesis},
  author={Yin, Tianwei and Gharbi, Micha{\"e}l and Park, Taesung and Zhang, Richard and Shechtman, Eli and Durand, Fr{\'e}do and Freeman, William T.},
  journal={arXiv preprint arXiv:2405.14867},
  year={2024}
}

@article{black2025rtc,
  title={Real-Time Execution of Action Chunking Flow Policies},
  author={Black, Kevin and Galliker, Manuel Y. and Levine, Sergey},
  journal={arXiv preprint arXiv:2506.07339},
  year={2025}
}

@article{wang2026remac,
  title={Real-Time Robot Execution with Masked Action Chunking},
  author={Wang, Haoxuan and Zhang, Gengyu and Yan, Yan and Shang, Yuzhang and Kompella, Ramana Rao and Liu, Gaowen},
  journal={arXiv preprint arXiv:2601.20130},
  year={2026}
}

@article{li2026flashvla,
  title={FlashVLA: Streaming Action Decoding for Fast and Asynchronous VLA Inference},
  author={Li, Zekai and Tang, Jiaming and Liu, Zhijian},
  journal={arXiv preprint arXiv:2608.27384},
  year={2026}
}

@article{motubrain2026asyncwam,
  title={World Action Models in Real Time: An Empirical Study of Smooth Execution via Asynchronous Deployment},
  author={{Motubrain Team}},
  journal={arXiv preprint arXiv:2608.01880},
  year={2026}
}

@inproceedings{sohl2015deep,
  title={Deep Unsupervised Learning using Nonequilibrium Thermodynamics},
  author={Sohl-Dickstein, Jascha and Weiss, Eric and Maheswaranathan, Niru and Ganguli, Surya},
  booktitle={ICML},
  pages={2256--2265},
  year={2015},
  organization={pmlr}
}

@article{ho2020denoising,
  title={Denoising Diffusion Probabilistic Models},
  author={Ho, Jonathan and Jain, Ajay and Abbeel, Pieter},
  journal={NeurIPS},
  volume={33},
  pages={6840--6851},
  year={2020}
}

@article{song2020score,
  title={Score-based Generative Modeling through Stochastic Differential Equations},
  author={Song, Yang and Sohl-Dickstein, Jascha and Kingma, Diederik P and Kumar, Abhishek and Ermon, Stefano and Poole, Ben},
  journal={arXiv preprint arXiv:2011.13456},
  year={2020}
}

@article{lipman2022flow,
  title={Flow Matching for Generative Modeling},
  author={Lipman, Yaron and Chen, Ricky TQ and Ben-Hamu, Heli and Nickel, Maximilian and Le, Matt},
  journal={arXiv preprint arXiv:2210.02747},
  year={2022}
}

@article{liu2022flow,
  title={Flow Straight and Fast: Learning to Generate and Transfer Data with Rectified Flow},
  author={Liu, Xingchao and Gong, Chengyue and Liu, Qiang},
  journal={arXiv preprint arXiv:2209.03003},
  year={2022}
}

@article{albergo2023stochastic,
  title={Stochastic Interpolants: A Unifying Framework for Flows and Diffusions},
  author={Albergo, Michael S and Boffi, Nicholas M and Vanden-Eijnden, Eric},
  journal={arXiv preprint arXiv:2303.08797},
  year={2023}
}

@article{salimans2022progressive,
  title={Progressive distillation for fast sampling of diffusion models},
  author={Salimans, Tim and Ho, Jonathan},
  journal={arXiv preprint arXiv:2202.00512},
  year={2022}
}

@inproceedings{sauer2024adversarial,
  title={Adversarial diffusion distillation},
  author={Sauer, Axel and Lorenz, Dominik and Blattmann, Andreas and Rombach, Robin},
  booktitle={European Conference on Computer Vision},
  pages={87--103},
  year={2024},
  organization={Springer}
}

@article{yang2026wam,
  title={WAM-OPD: On-Policy Distillation for World Action Models},
  author={Yang, Liuhaichen and Jiang, Zhuang and Sheng, Chenchao and Tang, Zezhi},
  journal={arXiv preprint arXiv:2608.22364},
  year={2026}
}

@article{black2025training,
  title={Training-time action conditioning for efficient real-time chunking},
  author={Black, Kevin and Ren, Allen Z and Equi, Michael and Levine, Sergey},
  journal={arXiv preprint arXiv:2512.05964},
  year={2025}
}

@article{tang2025vlash,
  title={Vlash: Real-time vlas via future-state-aware asynchronous inference},
  author={Tang, Jiaming and Sun, Yufei and Zhao, Yilong and Yang, Shang and Lin, Yujun and Zhang, Zhuoyang and Hou, James and Lu, Yao and Liu, Zhijian and Han, Song},
  journal={arXiv preprint arXiv:2512.01031},
  year={2025}
}

@article{cai2026aha,
  title={AHA-WAM: Asynchronous Horizon-Adaptive World-Action Modeling with Observation-Guided Context Routing},
  author={Cai, Jisong and Ling, Long and Chu, Shiwei and Liu, Zhongshan and Kang, Jiayue and Liang, Zhixuan and Xu, Wenjie and Mao, Yinan and Zhang, Weinan and Yang, Xiaokang and others},
  journal={arXiv preprint arXiv:2606.09811},
  year={2026}
}

@article{hirose2026asyncvla,
  title={Asyncvla: An asynchronous vla for fast and robust navigation on the edge},
  author={Hirose, Noriaki and Glossop, Catherine and Shah, Dhruv and Levine, Sergey},
  journal={arXiv preprint arXiv:2602.13476},
  year={2026}
}

@article{jiang2025asyncvla,
  title={Asyncvla: Asynchronous flow matching for vision-language-action models},
  author={Jiang, Yuhua and Cheng, Shuang and Ding, Yan and Gao, Feifei and Qi, Biqing},
  journal={arXiv preprint arXiv:2511.14148},
  year={2025}
}

@article{lu2026faster,
  title={Faster: Rethinking real-time flow vlas},
  author={Lu, Yuxiang and Liu, Zhe and Fan, Xianzhe and Yang, Zhenya and Hou, Jinghua and Li, Junyi and Ding, Kaixin and Zhao, Hengshuang},
  journal={arXiv preprint arXiv:2603.19199},
  year={2026}
}

@article{hu2021lora,
  title={Lora: Low-rank adaptation of large language models},
  author={Hu, Edward J and Shen, Yelong and Wallis, Phillip and Allen-Zhu, Zeyuan and Li, Yuanzhi and Wang, Shean and Wang, Lu and Chen, Weizhu},
  journal={arXiv preprint arXiv:2106.09685},
  year={2021}
}

@misc{lightx2v,
  author = {{LightX2V Contributors}},
  title = {{LightX2V}: Light Video Generation Inference Framework},
  year = {2025},
  publisher = {GitHub},
  journal = {GitHub repository},
  howpublished = {\url{https://github.com/ModelTC/lightx2v}}
}

@book{steele2004cauchy,
  title={The Cauchy-Schwarz master class: an introduction to the art of mathematical inequalities},
  author={Steele, J Michael},
  year={2004},
  publisher={Cambridge University Press}
}

@article{bu2025univla,
  title={Univla: Learning to act anywhere with task-centric latent actions},
  author={Bu, Qingwen and Yang, Yanting and Cai, Jisong and Gao, Shenyuan and Ren, Guanghui and Yao, Maoqing and Luo, Ping and Li, Hongyang},
  journal={arXiv preprint arXiv:2505.06111},
  year={2025}
}

@article{kim2025fine,
  title={Fine-tuning vision-language-action models: Optimizing speed and success},
  author={Kim, Moo Jin and Finn, Chelsea and Liang, Percy},
  journal={arXiv preprint arXiv:2502.19645},
  year={2025}
}

@article{pertsch2025fast,
  title={Fast: Efficient action tokenization for vision-language-action models},
  author={Pertsch, Karl and Stachowicz, Kyle and Ichter, Brian and Driess, Danny and Nair, Suraj and Vuong, Quan and Mees, Oier and Finn, Chelsea and Levine, Sergey},
  journal={arXiv preprint arXiv:2501.09747},
  year={2025}
}

@article{cen2025worldvla,
  title={Worldvla: Towards autoregressive action world model},
  author={Cen, Jun and Yu, Chaohui and Yuan, Hangjie and Jiang, Yuming and Huang, Siteng and Guo, Jiayan and Li, Xin and Song, Yibing and Luo, Hao and Wang, Fan and others},
  journal={arXiv preprint arXiv:2506.21539},
  year={2025}
}

@article{huang2026meanflownft,
  title={MeanFlowNFT: Bringing Forward-Process RL to Average-Velocity Generators},
  author={Huang, Yushi and Zhou, Xiangxin and Zhang, Jun and Bo, Liefeng and Pang, Tianyu},
  journal={arXiv preprint arXiv:2607.15273},
  year={2026}
}

@book{hairer1993solving,
  title={Solving ordinary differential equations I: Nonstiff problems},
  author={Hairer, E. and N{\o}rsett, S.P. and Wanner, G.},
  year={1993},
  publisher={Springer}
}

@inproceedings{gu2026anyflow,
  title={Anyflow: Any-step video diffusion model with on-policy flow map distillation},
  author={Gu, Yuchao and Fang, Guian and Jiang, Yuxin and Mao, Weijia and Han, Song and Cai, Han and Shou, Mike Zheng},
  booktitle={European Conference on Computer Vision},
  pages={163--181},
  year={2026},
  organization={Springer}
}
\bibliographystyle{iclr2027_conference}

\clearpage
\appendix

\addtocontents{toc}{\protect\setcounter{tocdepth}{2}}
\begingroup
    \renewcommand{\contentsname}{Appendix}
    \tableofcontents
\endgroup
\medskip

\numberwithin{equation}{section}
\numberwithin{figure}{section}
\numberwithin{table}{section}

\section{Limitations}
\label{app:limitation}

\textbf{1) Applicability.} RealtimeWAM targets MoT-based WAMs and is not directly applicable to shared-backbone architectures such as DreamZero~\citep{ye2026world}. Step distillation in shared-backbone WAMs more closely resembles step distillation in video generation, making action-specific characteristics harder to isolate and motivating our focus on MoT architectures. Within this scope, our framework supports both models that omit explicit future-video prediction at inference (\eg, Fast-WAM~\citep{yuan2026fast}) and those that retain it (\eg, Faster-WAM~\citep{zhao2026faster}).

\textbf{2) Training overhead.} The teacher-anchored loss $\mathcal{L}_{\mathrm{TA}}$ introduces additional training overhead by requiring multi-step rollouts from the frozen teacher. Nevertheless, this overhead remains manageable: as shown in Tab.~\ref{tab:training_overhead}, adding $\mathcal{L}_{\mathrm{TA}}$ increases training time from 5h~21min to 8h~3min (approximately 50\%), comparable to jointly fine-tuning the Video and Action Experts (8h~1min). By freezing the Video Expert, TACD reduces peak GPU memory from 65.98 to 24.73~GiB compared with joint fine-tuning, a 62.5\% reduction, while retaining the same memory footprint as action-only consistency distillation.

\begin{table}[!ht]
    \centering
    \setlength{\tabcolsep}{8pt}
    \renewcommand{\arraystretch}{1.1}
    \caption{Training time and peak GPU memory for 30{,}000 training iterations on 16 NVIDIA H100 GPUs. $\mathcal{L}_{\mathrm{CD}}$ and $\mathcal{L}_{\mathrm{TA}}$ denote the consistency and teacher-anchored losses, respectively. Video/Action indicate whether each expert is fine-tuned.}
    \label{tab:training_overhead}
\begin{tabular}{cccccc}
\toprule
$\mathcal{L}_{\mathrm{CD}}$ & $\mathcal{L}_{\mathrm{TA}}$
& Video & Action & Training Time$\downarrow$ & Peak Memory (GiB)$\downarrow$ \\
\midrule
\ding{51} & \ding{55} & \ding{51} & \ding{51} & 8h~1min & 65.98 \\
\ding{51} & \ding{55} & \ding{55} & \ding{51} & 5h~21min & 24.73 \\
\rowcolor{mycolor!30}
\ding{51} & \ding{51} & \ding{55} & \ding{51} & 8h~3min & 24.73 \\
\bottomrule
\end{tabular}

\end{table}

\clearpage
\section{Cross-Expert Wavefront Pipelining: Pseudocode}
\label{app:cewp_algorithm}

Our Cross-Expert Wavefront Pipelining (CEWP) supports MoT-based WAMs, including both models that omit explicit future-video prediction at inference (\eg, Fast-WAM~\citep{yuan2026fast}) and those that retain it (\eg, Faster-WAM~\citep{zhao2026faster}). Algorithm~\ref{alg:cewp} presents a unified schedule for one paired forward pass. Input preparation, attention masks, output heads, and denoising schedules follow the base model.

Let $\mathcal{C}$ denote the video-conditioned action blocks, indexed from $1$. Fast-WAM uses $\mathcal{C}=\{1,\ldots,N\}$, whereas our Faster-WAM implementation uses $\mathcal{C}=\{1,5,\ldots,N\}$. At each conditioning block, \texttt{Fuse} returns the single-block KV for Fast-WAM or the base model's weighted fusion of the accumulated interval KV for Faster-WAM. The video stream then records a readiness event, which the action stream waits on only before attention. Intervening action blocks require no cross-expert synchronization.

\begingroup
\definecolor{cewpcomment}{rgb}{0.25,0.5,0.5}
\definecolor{cewpfunc}{rgb}{0.85,0.18,0.50}
\lstset{
    language=Python,
    basicstyle=\fontsize{9}{10.8}\ttfamily\selectfont,
    keywordstyle=\color{blue}\bfseries,
    emph={P_v,P_a,Fuse,record,wait_event,Attn_v,Attn_a},
    emphstyle=\color{cewpfunc},
    commentstyle=\color{cewpcomment},
    stringstyle=\color{orange},
    columns=fullflexible,
    keepspaces=true,
    showstringspaces=false,
    breaklines=true,
    aboveskip=4pt,
    belowskip=2pt
}
\begin{algorithm}[H]
\caption{Cross-Expert Wavefront Pipelining for MoT-based WAMs.}
\label{alg:cewp}
\begin{lstlisting}
# hv: video states, including future tokens if used; ha: action states.
# P: Q/K/V projection, including key normalization and RoPE.
# R: remaining block operations; Mv, Ma: base-model masks.
def cewp(hv, ha, N, C, retain_video=False):
    caller = current_stream()
    Sv, Sa = Stream(), Stream()
    ready = {i: Event() for i in C}
    Sv.wait_stream(caller)
    Sa.wait_stream(caller)
    pending, cache = [], {}

    for i in range(1, N + 1):
        with stream(Sv):
            qv, kv, vv = P_v[i](hv)
            pending.append((kv, vv))
            if i in C:
                cache[i] = Fuse[i](pending)
                pending = []
                ready[i].record(Sv)  # Publish before video attention.
            hv = R_v[i](hv, Attn_v[i](qv, kv, vv, Mv[i]))

        with stream(Sa):
            qa, ka, va = P_a[i](ha)  # No video dependency yet.
            mask = None
            if i in C:
                Sa.wait_event(ready[i])  # Wait only at consumption.
                kc, vc = cache[i]
                kc.record_stream(Sa)
                vc.record_stream(Sa)
                ka, va = cat(kc, ka), cat(vc, va)
                mask = Ma[i]
            ha = R_a[i](ha, Attn_a[i](qa, ka, va, mask))

    caller.wait_stream(Sv)
    caller.wait_stream(Sa)
    ha.record_stream(caller)
    hv.record_stream(caller)
    action_pred = ActionHead(ha)
    video_pred = VideoHead(hv) if retain_video else None
    return action_pred, video_pred
\end{lstlisting}
\end{algorithm}
\endgroup
\clearpage

\section{Complete Error Derivation}
\label{app:error_derivation}

\subsection{Local Consistency and Endpoint Error}
\label{app:local_global_error}

We fix the model parameters and omit the shared conditioning $z(o,l)$. For $0<t\le1$, let $a_\tau^\star$ be the unique teacher ODE solution with $a_t^\star=a_t$, and let $a_0^\star$ denote its exact endpoint. All expectations below use the same training sampling distribution, and we assume finite second moments for the errors and velocity residuals.

Adding and subtracting the EMA prediction makes the two error terms explicit:
\begin{equation*}
    \begin{aligned}
        \mathbf e_{\mathrm{total}}
        &= f_{\theta_{\mathrm S}}(a_t,t)-a_0^\star \\
        &= \underbrace{f_{\theta_{\mathrm S}}(a_t,t)-f_{\theta_{\mathrm{ema}}}(\tilde a_s,s)}_{\mathbf e_{\mathrm{local}}}
        +\underbrace{f_{\theta_{\mathrm{ema}}}(\tilde a_s,s)-a_0^\star}_{\mathbf e_{\mathrm{global}}}.
    \end{aligned}
\end{equation*}
Expanding the squared norm and taking expectations gives
\begin{equation*}
    \mathbb E\|\mathbf e_{\mathrm{total}}\|_2^2
    =\mathbb E\|\mathbf e_{\mathrm{local}}\|_2^2
    +\mathbb E\|\mathbf e_{\mathrm{global}}\|_2^2
    +2\mathbb E\langle\mathbf e_{\mathrm{local}},\mathbf e_{\mathrm{global}}\rangle.
\end{equation*}
Cauchy--Schwarz~\citep{steele2004cauchy} bounds the cross term as
\begin{equation*}
    \begin{aligned}
        \left|\mathbb E\langle\mathbf e_{\mathrm{local}},\mathbf e_{\mathrm{global}}\rangle\right|
        &\le\mathbb E\!\left[\|\mathbf e_{\mathrm{local}}\|_2\|\mathbf e_{\mathrm{global}}\|_2\right] \\
        &\le\sqrt{\mathbb E\|\mathbf e_{\mathrm{local}}\|_2^2}
        \sqrt{\mathbb E\|\mathbf e_{\mathrm{global}}\|_2^2}.
    \end{aligned}
\end{equation*}
Substitution yields
\begin{equation}
    \begin{aligned}
        \mathbb E\|\mathbf e_{\mathrm{total}}\|_2^2
        &\le\left(\sqrt{\mathbb E\|\mathbf e_{\mathrm{local}}\|_2^2}
        +\sqrt{\mathbb E\|\mathbf e_{\mathrm{global}}\|_2^2}\right)^2, \\
        \mathbb E\|\mathbf e_{\mathrm{total}}\|_2^2
        &\ge\left(\sqrt{\mathbb E\|\mathbf e_{\mathrm{local}}\|_2^2}
        -\sqrt{\mathbb E\|\mathbf e_{\mathrm{global}}\|_2^2}\right)^2.
    \end{aligned}
    \label{eq:app_error_moments}
\end{equation}
Thus, a small expected local error alone does not ensure a small expected total error without control of the global target error. In particular, $\mathbb E\|\mathbf e_{\mathrm{local}}\|_2^2=0$ implies $\mathbb E\|\mathbf e_{\mathrm{total}}\|_2^2=\mathbb E\|\mathbf e_{\mathrm{global}}\|_2^2$: agreement with the EMA target does not by itself establish that the target is accurate.


\subsection{Equivalence to Weighted Endpoint Supervision}
\label{app:weighted_endpoint_supervision}

The student and teacher use the same noisy action $a_t$ and conditioning $z(o,l)$. By Eqs.~\eqref{eq:cd_prediction} and~\eqref{eq:teacher_rollout},
\begin{equation*}
    f_{\theta_{\mathrm S}}(a_t,t)=a_t-t\,v_{\theta_{\mathrm S}}(a_t,t),
    \qquad a_0^{\mathrm T}=a_t-t\,u_{\theta_{\mathrm T}}(a_t,t).
\end{equation*}
Subtracting the two endpoints gives
\begin{equation}
    f_{\theta_{\mathrm S}}(a_t,t)-a_0^{\mathrm T}
    =-t\!\left[v_{\theta_{\mathrm S}}(a_t,t)-u_{\theta_{\mathrm T}}(a_t,t)\right].
    \label{eq:app_endpoint_velocity}
\end{equation}
Since the teacher target is frozen, stop-gradient does not change its value. Taking squared norms, dividing by $t^2$, and averaging gives
\begin{equation}
    \begin{aligned}
        \mathcal L_{\mathrm{TA}}
        &=\mathbb E_{a_t,t}\!\left[\left\|v_{\theta_{\mathrm S}}(a_t,t)-u_{\theta_{\mathrm T}}(a_t,t)\right\|_2^2\right] \\
        &=\mathbb E_{a_t,t}\!\left[\left\|\frac{a_t-f_{\theta_{\mathrm S}}(a_t,t)}{t}-\frac{a_t-a_0^{\mathrm T}}{t}\right\|_2^2\right] \\
        &=\mathbb E_{a_t,t}\!\left[\frac{\left\|f_{\theta_{\mathrm S}}(a_t,t)-a_0^{\mathrm T}\right\|_2^2}{t^2}\right],\qquad t>0.
    \end{aligned}
    \label{eq:ta_endpoint_loss}
\end{equation}
Thus, the velocity loss is exactly a weighted endpoint loss, with $1/t^2$ inside the expectation. The equivalence holds for any teacher solver through the definition of $u_{\theta_{\mathrm T}}$.

\subsection{Total Endpoint Error Bound}
\label{app:teacher_anchoring_bound}

Let $\boldsymbol\delta=a_0^{\mathrm T}-a_0^\star$. Adding this numerical integration error to Eq.~\eqref{eq:app_endpoint_velocity} gives
\begin{equation*}
    \begin{aligned}
        \mathbf e_{\mathrm{total}}
        &=\left(f_{\theta_{\mathrm S}}(a_t,t)-a_0^{\mathrm T}\right)+\boldsymbol\delta \\
        &=-t\!\left[v_{\theta_{\mathrm S}}(a_t,t)-u_{\theta_{\mathrm T}}(a_t,t)\right]+\boldsymbol\delta,
    \end{aligned}
\end{equation*}
which recovers Eq.~\eqref{eq:ta_total_error}. Applying a consequence of Cauchy--Schwarz~\citep{steele2004cauchy}\footnote{For real vectors $p,q$, Cauchy--Schwarz gives $|\langle p,q\rangle|\le\|p\|_2\|q\|_2$. Together with $2ab\le a^2+b^2$, this yields $\|p+q\|_2^2=\|p\|_2^2+\|q\|_2^2+2\langle p,q\rangle\le2\|p\|_2^2+2\|q\|_2^2$.}, taking expectations, and using $0<t\le1$ yields
\begin{equation}
    \begin{aligned}
        \mathbb E\|\mathbf e_{\mathrm{total}}\|_2^2
        &\le2\mathbb E\!\left[t^2\left\|v_{\theta_{\mathrm S}}(a_t,t)-u_{\theta_{\mathrm T}}(a_t,t)\right\|_2^2\right]
        +2\mathbb E\|\boldsymbol\delta\|_2^2 \\
        &\le2\mathcal L_{\mathrm{TA}}+2\mathbb E\|\boldsymbol\delta\|_2^2.
    \end{aligned}
    \label{eq:app_ta_error_bound}
\end{equation}
This bounds the combined error $\mathbf e_{\mathrm{local}}+\mathbf e_{\mathrm{global}}$ and does not require the two components to decrease separately.

\paragraph{Teacher numerical integration error.}
Following the standard error analysis of the explicit Euler method~\citep{hairer1993solving}, we estimate the teacher's numerical integration error for a uniform $K$-step rollout. Let $h=t/K$, $\tau_k=t-kh$, and $\hat a_{\tau_k}$ denote the Euler iterates initialized at $\hat a_{\tau_0}=a_t$. Assume the teacher velocity is $L$-Lipschitz in state and $\|d^2a_\tau^\star/d\tau^2\|_2\le M$. For $D_k=\|\hat a_{\tau_k}-a_{\tau_k}^\star\|_2$, Taylor expansion gives $D_{k+1}\le(1+Lh)D_k+Mh^2/2$, with $D_0=0$. Summing this recurrence gives, for $L>0$,
\begin{equation*}
    \|\boldsymbol\delta\|_2=D_K
    \le\frac{Mh}{2L}\bigl[(1+Lh)^K-1\bigr]
    \le\frac{Me^{Lt}t^2}{2K}.
\end{equation*}
For $L=0$, the bound is $Mt^2/(2K)$. With uniform $L,M$, the numerical contribution to Eq.~\eqref{eq:app_ta_error_bound} is $O(K^{-2})$. Increasing the number of teacher rollout steps $K$ therefore tightens the error bound, providing explicit control over the teacher's numerical integration error.

\section{Inference Equivalence of Asynchronous Execution}
\label{app:inference_equivalence}

Cross-Expert Wavefront Pipelining (CEWP) changes only the execution schedule, using the same parameters, inputs, attention masks, denoising timestep, and operators as sequential inference. We fix the initial action noise and any future-video noise so that both executions start from identical states $(h_0^v,h_0^a)$. In our implementation, video KV cache entries are published after their required preprocessing, including key normalization and RoPE. The action stream waits on the corresponding event before concatenating video and action KV.

Suppose both executions receive identical states at block $i$. Equation~(\ref{eq:expert_projections}) then gives identical Q/K/V tensors. The event dependency in Eq.~(\ref{eq:wavefront_schedule}) ensures that action attention consumes the same completed video KV cache. Applying the same attention and remaining-block operators in Eq.~(\ref{eq:expert_block_updates}) therefore produces identical next-block states for both experts. Induction from $i=0$ gives, in exact arithmetic,
\begin{equation}
    \widehat{h}_i^v=h_i^v,
    \qquad
    \widehat{h}_i^a=h_i^a,
    \qquad i\in\{0,\ldots,N\}.
    \label{eq:cewp_hidden_equivalence}
\end{equation}
where the hat denotes asynchronous execution. The unchanged action output head preserves this equality. CEWP is thus another valid topological ordering of the original computation graph, requiring no architectural modification or additional training, and is independent of TACD.

For sparse conditioning in Algorithm~\ref{alg:cewp}, the same argument applies: each event is recorded after the required KV fusion, whose inputs and operator remain unchanged, while non-conditioning action blocks require no video KV. When future-video prediction is retained, the unchanged video output head likewise preserves the video prediction.

\section{Inference Latency Analysis}
\label{app:inference_latency}

We model the execution time of the two experts, excluding common costs such as VAE encoding. Let $T_v$ and $T_a$ denote the standalone latencies of one Video Expert pass and one Action Expert pass, respectively, and $S$ the number of action denoising steps. Since the video KV cache is computed once and reused across steps, sequential execution requires
\begin{equation}
    T_{\mathrm{seq}}(S)=T_v+S T_a.
    \label{eq:latency_sequential}
\end{equation}
After TACD, the Action Expert requires only one denoising step ($S=1$), reducing the sequential latency to
\begin{equation}
    T_{\mathrm{seq}}(1)=T_v+T_a.
    \label{eq:latency_one_step}
\end{equation}

CEWP further overlaps the two remaining expert passes through the block-wise schedule in Eq.~(\ref{eq:wavefront_schedule}). We model its latency as
\begin{equation}
    T_{\mathrm{pipe}}(1)=T_v+T_a-T_{\mathrm{overlap}}+\Delta,
    \label{eq:latency_pipeline}
\end{equation}
where $T_{\mathrm{overlap}}$ denotes the execution time saved through overlap, and $\Delta$ accounts for additional synchronization, scheduling, and resource-contention costs. Under ideal overlap, $T_{\mathrm{overlap}}\approx\min(T_v,T_a)$, giving
\begin{equation}
    T_{\mathrm{pipe}}(1)\approx\max(T_v,T_a)+\Delta.
    \label{eq:latency_ideal_overlap}
\end{equation}
This is an ideal reference: block dependencies and shared GPU resources limit the overlap attainable in practice. TACD and CEWP thus address complementary sources of latency: TACD reduces the number of action passes, while CEWP overlaps the remaining video and action passes.

\section{More Experimental Details and Results}
\label{app:experimental_details}

\subsection{Result sources and aggregation}
For results quoted from prior work in Tabs.~\ref{tab:main_results} and~\ref{tab:libero_plus}, we retain the scores and aggregates reported in Light-WAM~\citep{li2026light}, Flash-WAM~\citep{akbari2026flash}, and Faster-WAM~\citep{zhao2026faster}. On LIBERO and RoboTwin~2.0, we evaluate Fast-WAM and Faster-WAM using 10 action denoising steps, our implementations of the few-step distillation baselines (DMD, MeanFlow, and consistency distillation), and the RealtimeWAM variants. Unless otherwise specified, RealtimeWAM variants use the action-only LoRA post-training setting described in Sec.~\ref{lab:setup}, with the Video Expert kept frozen to maintain the teacher conditioning during distillation. For our evaluations, Overall is the arithmetic mean of the four suite-level success rates on LIBERO and of the Clean and Random success rates on RoboTwin~2.0. On LIBERO-Plus, Overall is the average success rate across perturbation categories, weighted by the number of trials in each category.
\label{app:test_intro}

\subsection{Efficient Kernel Implementation}
\label{app:efficient_kernels}

We apply five operator-level optimizations to reduce kernel launches and intermediate memory traffic. (1) We fuse modulation and residual gating into separate element-wise kernels, preserving BF16 intermediate rounding while keeping LayerNorm independent. (2) We fuse each Q/K RMSNorm into a single Triton kernel, retaining FP32 statistics and BF16 intermediate rounding. (3) We fuse RoPE type conversions and complex rotation into a single Triton kernel while retaining FP64 arithmetic. (4) We replace static all-True attention masks with \texttt{None}, preserving masks that exclude attention positions. (5) We pack self-attention Q/K/V and cross-attention K/V projections into single GEMMs, leaving cross-attention Q separate. These optimizations complement CUDA Graph, which reduces CPU submission overhead but does not automatically fuse operators within the graph.

\subsection{Detailed LIBERO Results}
\label{app:libero_results}

\begin{table*}[t]
    \centering
    \setlength{\tabcolsep}{3pt}
    \renewcommand{\arraystretch}{1.05}
    \caption{Detailed success rates (\%) on the four LIBERO task suites.}
    \label{tab:libero}
    \resizebox{\linewidth}{!}{
\begin{tabular}[t!]{l|cc|ccccc}
\toprule
\multirow{2}{*}{Method}
& \multicolumn{2}{c|}{NFE}
& \multicolumn{5}{c}{LIBERO} \\
& Video & Action
& Spatial$\uparrow$ & Object$\uparrow$ & Goal$\uparrow$
& LIBERO-10$\uparrow$ & Overall$\uparrow$ \\
\midrule
\multicolumn{8}{c}{\cellcolor[gray]{0.92}WAM and VLA Baselines} \\
\midrule
$\pi_{0}$~\citep{black2024pi_0}       & -- & -- & 96.8 & 98.8 & 95.8 & 85.2 & 94.1 \\
$\pi_{0.5}$~\citep{intelligence2025pi_}     & -- & -- & 98.8 & 98.2 & 98.0 & 92.4 & 96.9 \\
X-VLA~\citep{zheng2026x}           & -- & -- & 98.2 & 98.6 & 97.8 & 97.6 & 98.1 \\
Motus~\citep{bi2026motus}           & 10 & 10 & 96.8 & 99.8 & 96.6 & 97.6 & 97.7 \\
LingBot-VA~\citep{li2026causal}      & 20 & 50 & 98.5 & 99.6 & 97.2 & 98.5 & 98.5 \\
Fast-WAM~\citep{yuan2026fast}         & 1 & 10 & 97.0 & 99.4 & 96.6 & 94.8 & 97.0 \\
Faster-WAM~\citep{zhao2026faster}    & 1 & 10 & 99.2 & 99.6 & 98.4 & 98.4 & 98.9 \\
\midrule
\multicolumn{8}{c}{\cellcolor[gray]{0.92}Other Efficient Variants} \\
\midrule
Flash-WAM~\citep{akbari2026flash}       & 1 & 2 & 97.0 & 92.8 & 96.4 & 98.0 & 96.1 \\
Flash-WAM~\citep{akbari2026flash}       & 1 & 1 & 96.0 & 92.6 & 96.0 & 95.8 & 95.1 \\
Light-WAM~\citep{li2026light}       & 1 & 1 & 98.2 & 99.6 & 97.8 & 93.0 & 97.2 \\
\midrule
\multicolumn{8}{c}{\cellcolor[gray]{0.92}Few-Step Distillation} \\
\midrule
Fast-WAM~\citep{yuan2026fast}         & 1 & 1  & 94.8 & 96.8 & 96.6 & 91.8 & 95.0 \\
Mean Flow\textsuperscript{*}~\citep{geng2025meanflow} & 1 & 1 & 97.4 & 99.6 & 96.8 & 93.8 & 96.9 \\
Consistency Distillation\textsuperscript{*}~\citep{song2023consistency} & 1 & 1 & 96.4 & 99.4 & 96.8 & 95.2 & 96.9 \\
DMD\textsuperscript{*}~\citep{yin2024improved} & 1 & 1 & 97.0 & 99.4 & 95.8 & 92.2 & 96.1 \\
\rowcolor{mycolor!30}
RealtimeWAM\textsuperscript{*} & 1 & 1 & 97.0 & 99.4 & 97.2 & 94.2 & 97.0 \\
\rowcolor{mycolor!30}
RealtimeWAM\textsuperscript{\textdagger} & 1 & 1 & 98.8 & 99.8 & 99.6 & 97.6 & 99.0 \\
\bottomrule
\end{tabular}

    }
\end{table*}

Tab.~\ref{tab:libero} reports success rates on each LIBERO task suite, complementing the overall results in Tab.~\ref{tab:main_results}. With one-step action generation, RealtimeWAM\textsuperscript{*} matches the 97.0\% overall success rate of 10-step Fast-WAM, while RealtimeWAM\textsuperscript{\textdagger} achieves 99.0\%, comparable to 10-step Faster-WAM (98.9\%). Compared with Faster-WAM, RealtimeWAM\textsuperscript{\textdagger} improves Goal success by 1.2 percentage points (99.6\% vs.\ 98.4\%), with modest reductions on Spatial and LIBERO-10, indicating that overall performance is preserved despite variations across task suites.

\par\medskip
\noindent\begin{minipage}{\linewidth}
\subsection{Additional Ablation Study}
\label{app:teacher_interval}

\noindent\begin{minipage}[t]{0.48\linewidth}
    \vspace{0pt}
We further investigate the start and end points of the teacher rollout used to construct $\mathcal{L}_{\mathrm{TA}}$. We compare three intervals: $1\!\to\!0$, $t\!\to\!r$, and $t\!\to\!0$, where $1$ denotes pure noise, $0$ denotes the clean endpoint, and $t$ and $r<t$ denote intermediate noise levels. As shown in Tab.~\ref{tab:teacher_interval}, $t\!\to\!0$ achieves the highest overall success rate of 90.84\%, outperforming both the full-interval rollout $1\!\to\!0$ and the intermediate-interval rollout $t\!\to\!r$. We therefore adopt $t\!\to\!0$ as the default teacher rollout interval.
\end{minipage}\hfill
\begin{minipage}[t]{0.48\linewidth}
    \vspace{0pt}
    \centering
    \setlength{\tabcolsep}{8pt}
    \renewcommand{\arraystretch}{1.1}
    \setlength{\abovecaptionskip}{0pt}
    \captionof{table}{Effect of teacher rollout intervals on RoboTwin 2.0. Results are success rates (\%); best results are in bold.}
    \label{tab:teacher_interval}
    \resizebox{\linewidth}{!}{
\begin{tabular}{cccc}
\toprule
\makecell{Rollout\\Interval} & Clean$\uparrow$ & Random$\uparrow$ & Overall$\uparrow$ \\
\midrule
$1\!\to\!0$ & 91.26 & \textbf{89.90} & 90.63 \\
$t\!\to\!r$ & 90.96 & 89.46 & 90.23 \\
\rowcolor{mycolor!30}
$t\!\to\!0$ & \textbf{91.96} & 89.72 & \textbf{90.84} \\
\bottomrule
\end{tabular}
}
\end{minipage}
\end{minipage}
\par\medskip

\subsection{Latency on Additional GPUs}
\label{app:gpu_latency}

Tab.~\ref{tab:gpu_latency} extends the latency evaluation to RTX 5090 and RTX 4090D. TACD alone yields $3.5$--$5.0\times$ speedups, while CUDA Graph and CEWP further reduce latency across all four model--GPU configurations. With all optimizations enabled, RealtimeWAM achieves $14.1\times$ and $17.7\times$ speedups on Fast-WAM, and $6.9\times$ and $8.1\times$ on Faster-WAM, for RTX 5090 and RTX 4090D, respectively. The current kernels are optimized primarily for NVIDIA H-series GPUs (\eg, H100), which may explain their smaller gains on RTX 5090 and RTX 4090D. We expect dedicated kernel tuning for these GPUs to yield larger speedups.

\begin{table*}[t]
    \centering
    \setlength{\tabcolsep}{6pt}
    \renewcommand{\arraystretch}{1.1}
    \caption{Inference latency (ms) on additional GPUs. Each entry reports latency followed by the cumulative speedup in parentheses, relative to the unoptimized baseline for the same model and GPU.}
    \label{tab:gpu_latency}
    \resizebox{\linewidth}{!}{
\begin{tabular}{cccccccc}
\toprule
\multirow{2}{*}{TACD}
& \multirow{2}{*}{\makecell{CUDA\\Graph}}
& \multirow{2}{*}{CEWP}
& \multirow{2}{*}{\makecell{Efficient\\Kernels}}
& \multicolumn{2}{c}{Fast-WAM}
& \multicolumn{2}{c}{Faster-WAM} \\
\cmidrule(lr){5-6}\cmidrule(lr){7-8}
& & & & RTX 5090 & RTX 4090D & RTX 5090 & RTX 4090D \\
\midrule
\ding{55} & \ding{55} & \ding{55} & \ding{55} & 251.5 ($1.0\times$) & 480.8 ($1.0\times$) & 214.1 ($1.0\times$) & 362.4 ($1.0\times$) \\
\ding{55} & \ding{51} & \ding{55} & \ding{55} & 97.8 ($2.6\times$) & 106.1 ($4.5\times$) & 81.5 ($2.6\times$) & 96.0 ($3.8\times$) \\
\ding{51} & \ding{55} & \ding{55} & \ding{55} & 54.3 ($4.6\times$) & 96.8 ($5.0\times$) & 60.7 ($3.5\times$) & 91.7 ($4.0\times$) \\
\ding{51} & \ding{51} & \ding{55} & \ding{55} & 30.8 ($8.2\times$) & 35.3 ($13.6\times$) & 38.5 ($5.6\times$) & 54.9 ($6.6\times$) \\
\ding{51} & \ding{51} & \ding{51} & \ding{55} & 24.3 ($10.3\times$) & 31.2 ($15.4\times$) & 34.7 ($6.2\times$) & 51.1 ($7.1\times$) \\
\rowcolor{mycolor!30}
\ding{51} & \ding{51} & \ding{51} & \ding{51} & 17.8 ($14.1\times$) & 27.2 ($17.7\times$) & 30.9 ($6.9\times$) & 44.7 ($8.1\times$) \\
\bottomrule
\end{tabular}
    }
\end{table*}

\begin{figure}[!t]
    \centering
    \includegraphics[width=\linewidth]{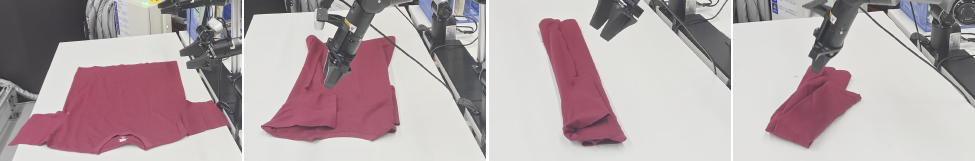}
    \caption{\textbf{Real-world garment folding with AgileX PiPER.} Four keyframes from the recorded rollout show the initial unfolded T-shirt, the configuration after the first fold, a narrow configuration after further lengthwise folding, and the final folded configuration. The rollout was executed with the one-step RealtimeWAM\textsuperscript{*} model. Consecutive actions proceed without perceptible pauses, demonstrating the smooth, low-latency inference that sustains long-horizon deformable-object manipulation.}
    \label{fig:garment_folding_sequence}
\end{figure}

\subsection{Real-World Evaluation of Long-Horizon Garment Folding}
\label{app:garment_folding}
We qualitatively evaluate RealtimeWAM\textsuperscript{*} on a real-world T-shirt-folding task using a bimanual setup comprising two PiPER robotic arms from AgileX Robotics\footnote{\url{https://global.agilex.ai/products/piper}}, with policy inference running on an NVIDIA RTX 4090 GPU. As shown in Figure~\ref{fig:garment_folding_sequence}, the task involves successive grasping and folding operations that transform an unfolded garment into a compact configuration. It requires temporally coordinated decisions because each fold changes the garment geometry and the grasp locations available for subsequent actions. Early misalignment can therefore affect later folds, making this task a useful setting for examining long-horizon manipulation of deformable objects. 

To meet the real-time requirement of this sequential setting, we deploy the RealtimeWAM\textsuperscript{*} model distilled from Fast-WAM, using TACD for one-step action generation and CEWP for block-wise overlap between the Video and Action Experts. In the physical evaluation, consecutive actions proceeded without perceptible pauses, enabling smooth real-time execution even without execution-level asynchrony between policy inference and robot execution, such as Real-Time Chunking (RTC)~\citep{black2025rtc}. The resulting low-latency inference allows the policy to maintain continuous control throughout the long-horizon folding sequence. In the context of RealtimeWAM\textsuperscript{*}, this task motivates preserving action quality under one-step action generation while reducing inference overhead during sequential execution.

\end{document}